%% file: main.tex
\documentclass[runningheads]{llncs}

\usepackage{eccv}

\input{preamble}

\begin{document}

\title{Decompose, Compare, and Decide: \\ Multimodal LLMs are Implicit Few-Shot Learners}

\titlerunning{DeCoDe: Multimodal LLMs are Implicit Few-Shot Learners}


\author{Yunhan Wang \and Eshika Khandelwal \and Edson Araujo \and Walid Bousselham \and \\ Nina Shvetsova \and Hilde Kuehne \\
}

\authorrunning{Y.~Wang et al.}

\institute{Tuebingen AI Center, University of Tuebingen, Germany \\
\email{yunhan.wang@student.uni-tuebingen.de}}


\maketitle

\input{ECCV_2026/sec/0_abstract}
\input{ECCV_2026/sec/1_intro}
\input{ECCV_2026/sec/2_relwork}

\input{ECCV_2026/sec/3_method}
\input{ECCV_2026/sec/4_evaluation}

\input{ECCV_2026/sec/5_conclusion}

%
%

\bibliographystyle{splncs04}
\bibliography{ECCV_2026/bib/longstrings,ECCV_2026/bib/main}

\clearpage
\appendix
\FloatBarrier
\input{ECCV_2026/sec/6_supplement}

\end{document}

%% file: preamble.tex
\usepackage{eccvabbrv}

\usepackage{graphicx}
\usepackage{booktabs}

\usepackage[accsupp]{axessibility}  

\usepackage{hyperref}

\usepackage{orcidlink}

\usepackage{multirow}
\usepackage{booktabs}
\usepackage{tabularx}
\usepackage{array}
\usepackage{colortbl,xcolor,adjustbox}
\usepackage{float}
\usepackage{wrapfig}

\usepackage{amssymb}
\usepackage{pifont}
\usepackage{soul}
\usepackage{placeins}
\usepackage{makecell}

\definecolor{stdblue}{RGB}{207,222,243}      
\definecolor{novelamber}{RGB}{252,228,190}   

\definecolor{secblue}{RGB}{232,240,252}      
\definecolor{secorange}{RGB}{255,243,224}    
\definecolor{gold}{RGB}{255,215,0}           
\definecolor{silver}{RGB}{210,210,210}       
\definecolor{rankone}{RGB}{180,0,0}          
\definecolor{ranktwo}{RGB}{0,100,160}        
\definecolor{modeltint}{RGB}{245,245,245}

\definecolor{woodbrown}{RGB}{92,51,23}

\definecolor{almond}{rgb}{0.98, 0.95, 0.93} 
\definecolor{lightgreenrow}{rgb}{0.92, 0.97, 0.92}
\definecolor{lightredrow}{rgb}{0.98, 0.92, 0.92}

\newcommand{\xmark}{\ding{55}} 
\newcommand{\cmark}{\ding{51}} 

\usepackage[most]{tcolorbox}
\newtcolorbox{promptbox}[1][]{%
  enhanced,
  boxrule=0.4pt,
  colback=black!2,
  colframe=black!25,
  arc=2pt,
  left=6pt,right=6pt,top=6pt,bottom=6pt,
  fontupper=\footnotesize\ttfamily,
  #1
}

\usepackage{caption}

%% file: ECCV_2026/sec/0_abstract.tex
\begin{abstract}

Multimodal Large Language Models (MLLMs) have demonstrated remarkable abilities when analyzing images, yet translating these capabilities to few-shot image classification remains challenging. To bridge this gap, we present DeCoDe, a simple yet effective technique that enables off-the-shelf MLLMs to act as strong few-shot classifiers without any additional training.
Our approach builds on the idea of few-shot classification as a set of pairwise image comparisons, decomposing the task into a set of binary decisions. 
Given a query image and a support image from a candidate class, the MLLM is prompted to decide whether the two images depict the same class.
The logit corresponding to an affirmative response is then used as a similarity score to assign the query image to the most likely class.
While this already yields good results, we show that providing additional high-level information, such as the data domain, to the model further improves performance.
Our evaluation provides an extensive analysis of various inference variants on a suite of twelve datasets, six established and six newly curated few-shot benchmarks spanning across diverse domains. The results show that the proposed simple decomposition technique can turn off-the-shelf MLLMs into powerful few-shot learners, significantly outperforming current state-of-the-art few-shot methods on both standard and novel domains. Code is available at \url{https://github.com/yunhanwang1105/DeCoDe}.

\keywords{Few-Shot Learning \and Multimodal Large Language Models \and Vision-Language Models \and In-Context Learning}

\end{abstract}

%% file: ECCV_2026/sec/1_intro.tex
\section{Introduction}
\label{sec:intro}

\begin{figure}[t]
\centering
\includegraphics[width=\linewidth]{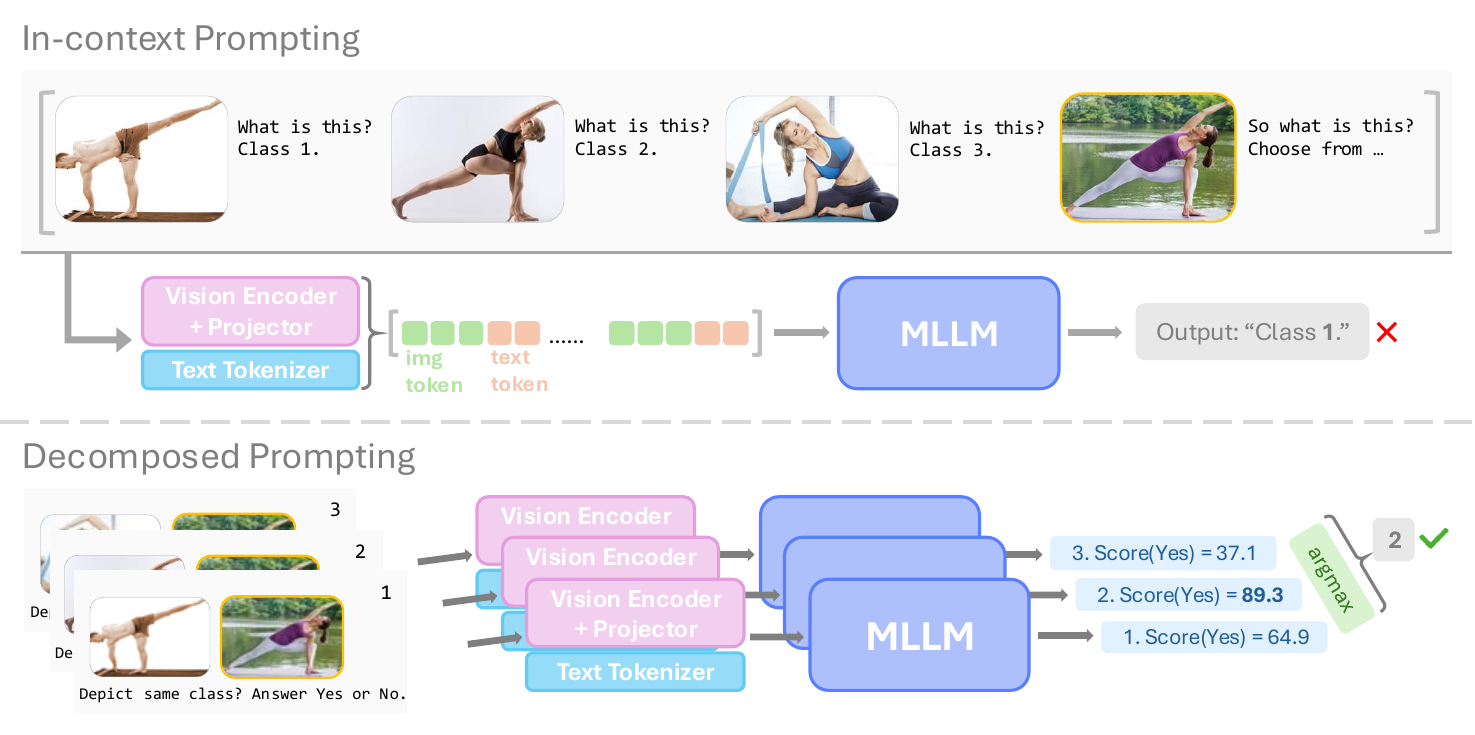}
\caption{
We propose a decomposed prompting technique (DeCoDe) for few-shot classification with MLLMs. We \textbf{decompose} the task into pairwise support–query comparisons, asking whether two images belong to the same class. 
By ranking the model’s affirmative responses across candidate pairs (\textbf{compare}) and selecting the highest-scoring logit as the predicted class (\textbf{decide}), MLLMs become strong few-shot classifiers without any training. Unlike standard in-context prompting, which relies heavily on semantic label names, our decomposed approach succeeds even with anonymized labels by forcing the model to perform direct visual-to-visual comparison. 
}
\label{fig:teaser}
\end{figure}


Few-shot learning (FSL) enables models to generalize to new tasks and domains using only a small number of labeled examples per class. This can be especially valuable for real-world scenarios where large annotated datasets are unavailable or adaptation to novel tasks is required at inference time. Early few-shot methods required model training on a set of base classes in order to enable recognition of held-out novel classes~\cite{vinyals2016matching, snell2017prototypical, finn2017model, sung2018learning, ravi2017optimization, Kukleva_2021_ICCV}. With the emergence of foundation models trained on large-scale web data via self- or weakly-supervised objectives, such as DINO~\cite{oquab2023dinov2} or CLIP~\cite{lpclip}, the focus shifted towards exploring how pretrained representations can be leveraged for improved few-shot classification with minimal or no additional training~\cite{chi2025learning, proker, tipadapter, coop, coopop, clipadapter, tian2020rethinking}. More recently, multimodal large language models (MLLMs), such as LLaVA-OneVision~\cite{li2024llava}, InternVL3~\cite{zhu2025internvl3}, or Qwen3-VL~\cite{bai2025qwen3}, have demonstrated strong vision-language capabilities, raising the question of whether such models can also serve as effective few-shot learners~\cite{liu2024, Yang_2025_Verbalized}.





So far, MLLMs have been explored for few-shot learning in three ways: by extracting discriminative embeddings from the internal representations of the MLLM~\cite{sav}, 
by verbalizing discriminative visual features and matching them to query images~\cite{Yang_2025_Verbalized}, or via in-context learning where support and query images are jointly provided in a single prompt~\cite{liu2024}. However, in-context prompting has been shown to perform poorly for few-shot image classification when used off-the-shelf~\cite{sav,liu2024}, often requiring additional fine-tuning to encourage stronger reliance on the support examples~\cite{liu2024}. 

To address this limitation, the proposed approach reformulates few-shot classification as a set of pairwise comparisons. Instead of reasoning over all support examples jointly, the few-shot task is decomposed into binary prompts asking the MLLM whether a query image and a support image belong to the same class.
Unlike existing methods, particularly~\cite{liu2024,Yang_2025_Verbalized}, the proposed technique considers the logit of the \texttt{Yes} token to rank all candidate pairs and assign the query image to the most likely class.
While this simple technique, when applied to standard MLLMs, is already sufficient to outperform current state-of-the-art methods, it shows that the models can benefit from contextual information, such as the domain of the data. 
Namely, we found that the performance of the pairwise decomposed setting further improves when models are given the topic of the data, usually based on the dataset name, such as cars, yoga poses, etc. 
We thus coin the respective workflow as \textit{\textbf{De}compose}, \textit{\textbf{Co}mpare}, \textit{\textbf{De}cide} (DeCoDe) and show that this technique can transform state-of-the-art MLLMs into high-performing few-shot learners.

However, evaluating the few-shot capabilities of web-scale pretrained models remains challenging. In particular, the performance of models on standard few-shot benchmarks may be influenced by potential overlap between benchmark datasets and the models’ pretraining corpora. As a result, models may rely on memorized knowledge rather than adapting to the provided support examples~\cite{kravets2025rethinking}.
To address this issue, we conduct an extensive analysis on two sets of datasets and in two evaluation modes.  \textit{First}, with respect to datasets, we consider six standard few-shot datasets, as well as six novel datasets, which can be considered as out-of-domain for web-scale trained models. 
We observe that most MLLMs already achieve near-perfect accuracy on the standard datasets in a 0-shot setting, often even exceeding 1-shot performance. In contrast, we consider novel, out-of-domain datasets as datasets that show substantially low 0-shot performance as well as a clear improvement from 0-shot to a simple 1-shot baseline. We consider this as an indication that the model might have seen less of the original data and needs to rely more on visual cues.
\textit{Second}, we consider two different few-shot settings for MLLMs: first, we consider a semantic setting, in which semantically meaningful class labels are provided together with the support images, and, second, we consider an anonymized setting, where support images are labeled only with numerical identifiers (e.g., \texttt{Class 1}, \texttt{Class 2}), preventing the model from relying directly on textual knowledge.
Our analysis shows that the anonymization alone is enough to reduce the performance of MLLMs for in-context learning, even on standard benchmarks, whereas the proposed decomposition is able to significantly increase the performance in all cases, with and without class labels, as well as for standard and novel datasets.  

%
The contributions of this work can be summarized as follows: 
\begin{itemize}
    \item We propose a simple, yet effective technique of \textit{Decompose, Compare, Decide} (DeCoDe), to turn MLLMs into powerful few-shot learners by ranking the logits corresponding to an affirmative response to a binary prompt question. 
    \item We show that decomposed MLLMs can further benefit from high-level cues such as domain information and propose a new few-shot evaluation scenario by providing the global dataset domain as context. 
    \item We conduct a systematic evaluation of current MLLMs for few-shot learning across standard and novel out-of-domain datasets and under different modes, including anonymized class labels, revealing when models rely on memorized pretraining knowledge versus true few-shot adaptation.
\end{itemize}


%% file: ECCV_2026/sec/2_relwork.tex
\section{Related Works}
\label{sec:relwork}



\paragraph{VLM Embeddings for Few-Shot Learning.}

Vision-Language Models (VLMs) such as CLIP provide a strong foundation for FSL by enabling classification based on embedding similarity.
Prior work has adapted such embeddings for few-shot tasks via prompt learning~\cite{coop, coopop, prograd, kgcoop, plot, maple} or adapter-based methods~\cite{clipadapter, taskres, susx, tipadapter}. 
More recent works explore additional adaptation strategies. Two-Stage Few-Shot (2SFS)~\cite{2SFS} performs two-stage PEFT-based representation learning; Logit Deconfusion ~\cite{ldc} addresses class bias at the logit level; ProKeR~\cite{proker} applies training-free kernel regression over pretrained CLIP features; LoRA Recycle~\cite{recyclelora} enables tuning-free adaptation via reuse of pre-trained LoRAs \cite{hu2022lora}. CAML~\cite{fifty2023context} is a large-scale, meta-trained framework built on a ViT \cite{dosovitskiy2020image} backbone that enables in-context classification using frozen image embeddings. 
Unlike methods that modify pretrained encoders, our proposed technique does not adapt the underlying representation or update parameters. Instead, we investigate how frozen MLLMs can perform few-shot classification purely through structured inference.

Finally, Kravets et al.~\cite{kravets2025rethinking} discuss the problem that standard CLIP few-shot evaluations are partially transductive due to pretraining overlap, and introduce an unlearning-based pipeline to construct inductive benchmarks that reveal substantial performance drops in existing methods.
We follow this idea, but instead of unlearning and thus changing the model parameters, we use alternative datasets that can be considered more out-of-domain for web-trained multimodal models. 

\paragraph{MLLMs for Few-Shot Learning.}

Beyond adapting pretrained VLM encoders, SAVs~\cite{sav} extracts representations from MLLMs for downstream few-shot classification. VRL~\cite{Yang_2025_Verbalized} leverages MLLMs to automatically derive interpretable verbalized features that capture inter-class differences and intra-class commonalities. 
Other works explore generative few-shot prompting with MLLMs~\cite{liu2024, sav}, yet such models often remain less effective with respect to embedding-based few-shot methods, despite strong performance on open-ended vision–language tasks. 
A central challenge is their reliance on semantic language priors rather than visual evidence from the support set~\cite{baldassini2024}. GFSL~\cite{liu2024} addresses this in part through label anonymization, applied within a fine-tuning strategy. Across these works, evaluations vary considerably in scope: SAVs focus on standard benchmarks, VRL reports results on novel out-of-domain datasets. Building on this idea, we jointly examine zero-shot and few-shot regimes, the effect of semantic versus anonymized labels, and evaluate across both standard and novel datasets, providing a unified evaluation across these axes. 


%% file: ECCV_2026/sec/3_method.tex
\section{Method}
\label{sec:method}
In this section, we introduce our inference framework for few-shot classification. 
~\cref{fig:teaser} illustrates the inference pipeline of our framework.
Given multimodal context consisting of images and text, the MLLM is prompted to predict the class of a query image based on the provided support examples.
Depending on the prompting paradigm, the prediction is either obtained directly from the model's generated response or derived from the model's output logits for specific answer tokens (e.g., `Yes').


~\Cref{tab:variants} summarizes the inference variants evaluated in this work. 
The variants differ along three dimensions: 
(i) the prompting paradigm (in-context learning vs.\ decomposed pairwise prompting);
(ii) the use of dataset-level domain information; and
(iii) the anonymization of semantic class labels.
In addition, we evaluate a 0-shot setting as a baseline to assess whether the models can classify a query based solely on its prior knowledge without visual support examples.
In all experiments, we follow the prompt design of~\cite{liu2024} and explore more prompt options in the supplementary material. 

\subsection{Problem Formulation}
\label{subsec:problem_formulation}

Classical few-shot learning protocols~\cite{vinyals2016matching, snell2017prototypical, finn2017model} divide a dataset into 
a set of base classes used for supervised pre-training and a set of novel classes reserved for few-shot evaluation.
In contrast, training-free 
approaches 
leverage
VLMs and MLLMs~\cite{tipadapter, susx, sav, proker}, bypassing supervised base-class training, thanks to rich representations acquired during large-scale web pretraining.
In this paper, we follow a training-free protocol in the standard \textit{N-way K-shot} setting. The model's task is to classify a query image into one of $N$ classes given a support set of $K$ labeled examples per class. We define the support set $\mathcal{S}$ to be sampled from the test set as:  
\begin{equation}
\mathcal{S} = \bigl\{(x_{n,k}^s,\, c_n) \mid n=1,\dots,N,\; k=1,\dots,K\bigr\},
\end{equation}
where $N$ is the number of classes sampled, $K$ is the number of samples per class, and  $\mathcal{C} = \{c_1, \dots, c_N\}$ is the set of corresponding class names where each $c_n$ is the semantic name of the $n$-th class.
We formulate the query set~$\mathcal{Q}$ as:
\begin{equation}
\mathcal{Q} = \{(x^q, c_y)\},
\end{equation}
where $x^q$ is the query image and $y \in \{1, \dots, N\}$ is its ground-truth class index with respect to the support classes $\mathcal{C}$.

\input{ECCV_2026/tab/variants}

\begin{figure}[t]
\centering
\includegraphics[width=\linewidth]{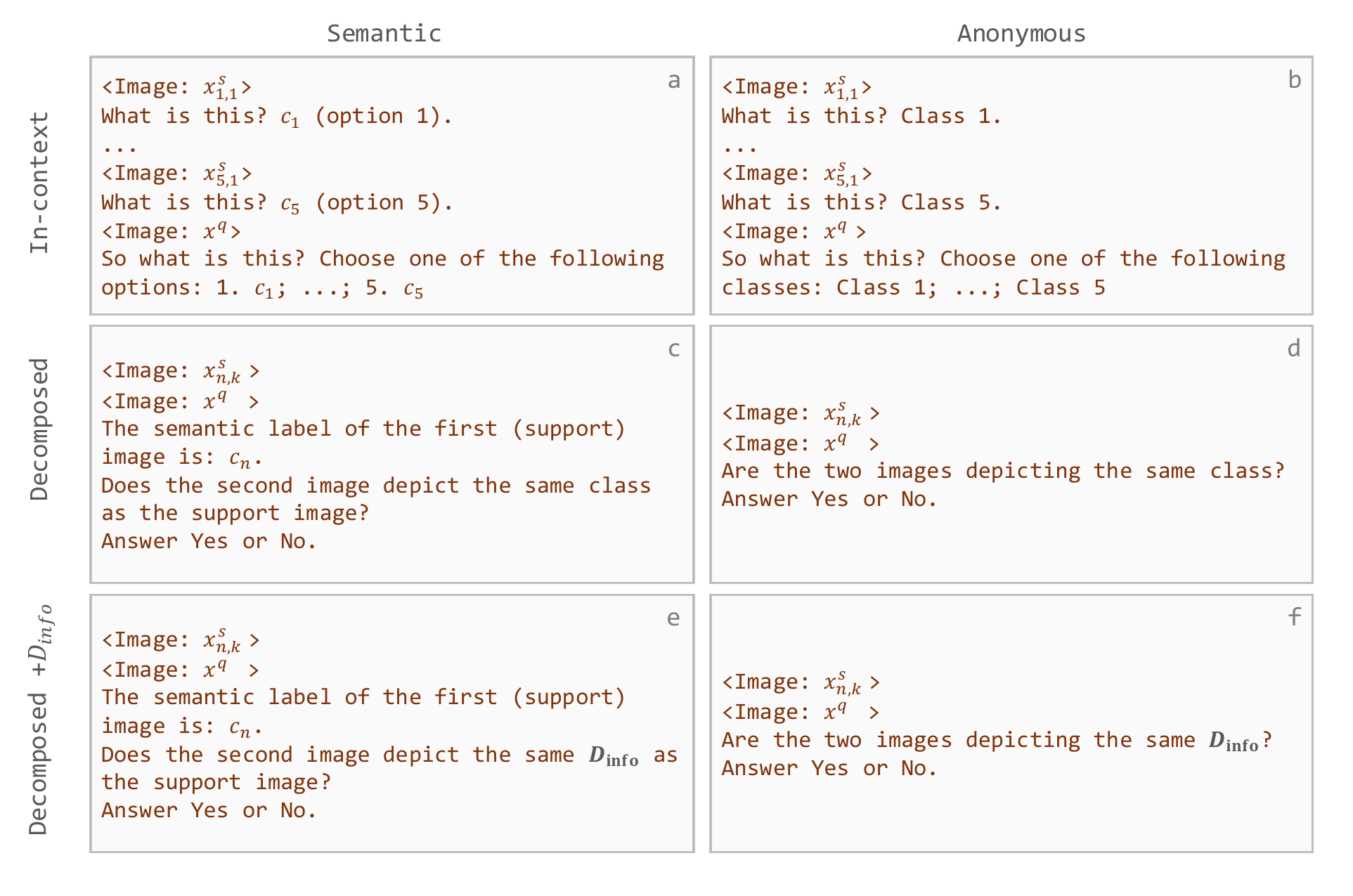}
\caption{
Variants of prompt formulations used in our experiments. 
\textbf{(a)} standard in-context prompting with semantic labels. 
\textbf{(b)} standard in-context prompting with anonymized labels. 
\textbf{(c)} decomposed pairwise prompting with semantic labels. 
\textbf{(d)} decomposed pairwise prompting with anonymized labels.
\textbf{(e)} decomposed pairwise prompting with domain information and semantic labels. 
\textbf{(f)} decomposed pairwise prompting with domain information and anonymized labels.
}
\label{fig:prompts}
\end{figure}

\subsection{In-Context Few-Shot Classification with MLLMs }
\label{subsec:incontext}
To assess the few-shot capabilities of current MLLMs, we consider few-shot learning as an in-context learning task.
Given a support set $\mathcal{S}$ and a query image $x^q$, the model is prompted to predict the class of the query image via multiple-choice selection over the candidate classes.
The prompt consists of the concatenation of all support images with their corresponding class labels, followed by the query image and the list of candidate class options.
The resulting prompt $p$ is structured as: 
\begin{equation}
p = x^s_{1,1} \oplus \texttt{Str} \oplus c_1 \dots x^s_{N,K} \oplus \texttt{Str} \oplus c_N\ \oplus  \\
x^q \oplus \texttt{Str} \oplus 1. \, c_1; 2. \, c_2; \dots ;  \\ 
\end{equation}


\noindent The model is expected to output a class index $\hat{y} \in \{1,\dots,N\}$, which we extract from the generated text via regular expression matching. Full prompt can be found in the top-left of Figure \ref{fig:prompts} (a). We further instruct the model to output the expected answer format. Exact prompts and instruction-following errors are reported in the supplementary material.
%

\subsection{Decomposed Few-Shot Classification with MLLMs }
\label{subsec:decomposition}
In contrast to in-context learning, we propose to decompose the few-shot classification problem into a set of pairwise comparisons. For each support example and the query image, the model is prompted to determine whether the two images belong to the same class. Given $(\mathcal{S}, x^q)$, we decompose the support set into a set of pairwise prompts: 
\begin{equation}
\mathcal{P} = \bigl\{p_{n,k} \mid n=1,\dots,N,\; k=1,\dots,K\bigr\},
\end{equation}
\noindent Each prompt compares the query image with a support example:
\begin{equation}
p_{n,k} =  x^s_{n,k} \oplus x^q \oplus \texttt{Str} \oplus c_n  \\ 
\end{equation}


All prompts are independently processed by the model. The full prompt can be found in the mid-right of Figure \ref{fig:prompts} (c). For each prompt $p_{n,k}$, the model's output logit for the token \texttt{Yes} is used as the score indicating whether the pair belongs to the same class:
\begin{equation}
\label{eq:pairwise}
s_{n,k}^{\mathrm{pair}} = logit(\texttt{Yes} \mid p_{n,k}),
\qquad
\hat{y} = \arg\max_{n}\; \frac{1}{K}\sum_{k=1}^{K} s_{n,k}^{\mathrm{pair}},
\end{equation}
where $\hat{y} \in \{1, \dots, N\}$ is the index with the highest averaged pairwise score as the predicted class. This choice is consistent with recent VQA-based scoring approaches such as VQAScore~\cite{lin2024evaluating}, which scores image-text alignment using the \texttt{Yes} token for evaluating text-to-visual generation; in contrast, we score support--query image-pair matches for few-shot image classification.

\subsection{Decomposed Prompting with Domain Information}
\label{subsec:domain}
\input{ECCV_2026/tab/Dinfo_table_main_text}
Decomposed prompting resembles the classical few-shot classification scenario, where a single representation is compared against a set of others. However, MLLMs also offer the ability to further refine this setting by encouraging the model to focus on dataset-specific concepts.
To leverage those capabilities, we replace the generic term \texttt{class} with a domain-appropriate term $D_{\text{info}}$ (e.g., \texttt{pose}, \texttt{texture}, \texttt{species})  
that reflects the concept represented in a given dataset. The full prompt can be found in the bottom-right of Figure \ref{fig:prompts} (e).  The descriptors $D_{\text{info}}$ are inferred for each dataset as shown in Table \ref{tab:dinfo_mapping_main_text}.
This grounds each comparison in both visual and textual semantics, allowing the MLLM to leverage its pretrained 
domain-level
knowledge alongside the visual evidence from the support image\footnote{For mini-ImageNet and DomainNet, we use a generic instruction asking the model to focus on the depicted concept rather than domain or texture. See supplement.}. We further provide preliminary evidence in the supplementary material that $D_{\text{info}}$ can be inferred automatically with competitive performance, and we also include example dataset labels to illustrate how $D_{\text{info}}$ captures domain-level information distinct from class labels.

\subsection{Prompting Without Class Labels }
\label{subsec:labels}

To isolate the model's ability to learn from visual examples alone, we replace all semantic class names $c_n$ with abstract identifiers
(e.g., \texttt{Class~1}, \dots, \texttt{Class~N}) in both the support demonstrations and the option list.
All other formatting stays identical.
This forces the MLLM to rely on visual similarity rather than leveraging any prior knowledge associated with class names. The full anonymous prompts can be found in the right column of Figure \ref{fig:prompts}.

\subsection{Episode Construction}
\label{subsec:episodes}
We construct evaluation episodes using the standard $N$-way $K$-shot protocol~\cite{tipadapter, susx, sav, proker}. 
In each episode $e$, an $N$-way label set is sampled, $\mathcal{C} = \{c_1, \dots, c_N\}$, with support set
$\mathcal{S} = \{(x_{n,k}^s, c_n)\}_{n=1,k=1}^{N \times K}$ containing $K$ examples per class, and a single query sample
$\mathcal{Q} = \{(x^q, c_y)\}$.
Given $(\mathcal{S}, x^q)$, the model predicts $\hat{y} \in \{1, \dots, N\}$.
Performance is measured as mean episodic accuracy over a set of evaluation episodes $\mathcal{E}$:
\[
\mathrm{Acc}
=
\frac{1}{|\mathcal{E}|}
\sum_{e \in \mathcal{E}}
\mathbf{1}[\hat{y}_e = y_e].
\]
For each dataset, we pre-sample a fixed set of 1000 evaluation episodes using a fixed random seed. 
We verify that cumulative accuracy stabilizes well within 1000 episodes. 
All evaluation runs are performed on the same set of episodes to ensure fair comparison across models and variants.


%% file: ECCV_2026/tab/variants.tex
\begin{table}[t]
\centering
\resizebox{\linewidth}{!}{%
\begin{tabular}{l|l|l|c|l}
  \toprule
  \textbf{Variant} & \textbf{Paradigm} & \textbf{Label} & \textbf{Domain} & \textbf{Description} \\
  \midrule
  0-shot w. labels           & In-context  & Semantic & \xmark & No support images; only query image and labels given \\
  1-shot w. labels           & In-context  & Semantic & \xmark & Support images with labels given \\
  1-shot w/o labels          & In-context  & Anonymous & \xmark &  Support images with anonymized labels e.g., \texttt{Class~1}, \dots, \texttt{Class~5}  \\
  \midrule
  \rowcolor{almond}\multicolumn{5}{l}{\textit{Ours (Decomposition)}} \\[-1pt]

  1-shot dec. w. labels              & Decompose  & Semantic  & \xmark &  Binary comparisons w. label added to binary question \\
  1-shot dec.+$D_{\text{info}}$ w. labels   & Decompose  & Semantic  & \cmark &  Binary comparisons w. label + Domain-adapted prompt wording \\
  1-shot dec. w/o labels             & Decompose  & Anonymous & \xmark &  Binary comparisons w/o label\\
  1-shot dec.+$D_{\text{info}}$ w/o labels  & Decompose  & Anonymous & \cmark &  Binary comparisons w/o label + Domain-adapted prompt wording \\
  \bottomrule
\end{tabular}%
}
\vspace{2mm}
\caption{Overview of inference variants. Each variant is characterized by its task structure, label naming strategy, and distinguishing mechanism. \textit{dec.}~=~decompose, $D_{\text{info}}$ = domain information.}
\label{tab:variants}
\end{table}

%% file: ECCV_2026/tab/Dinfo_table_main_text.tex
%

\begin{wraptable}{r}{0.47\linewidth}
\vspace{-3mm}
\centering
\tiny
\setlength{\tabcolsep}{1.2pt}
\renewcommand{\arraystretch}{0.78}

\caption{Mapping of datasets to their domain information text.}
\label{tab:dinfo_mapping_main_text}
\vspace{-1mm}

\begin{tabular}{@{}p{0.56\linewidth}p{0.38\linewidth}@{}}
\toprule
\textbf{Dataset} & \textbf{$D_{\text{info}}$ text} \\
\midrule
mini-ImageNet~\cite{vinyals2016matching} & -- \\
UCF101~\cite{soomro2012ucf101} & action category \\
CUB~\cite{wah2011caltech} & bird species \\
Aircraft~\cite{maji13fine-grained} & aircraft variant \\
Dogs~\cite{khosla2011novel} & dog breed \\
DomainNet~\cite{Peng_2019_ICCV} & -- \\
Lego~\cite{kaggle_lego} & Lego brick type \\
Industrial~\cite{kaggle_industrial} & industrial product \\
Yoga~\cite{kaggle_yoga} & yoga pose \\
Egyptian hieroglyph~\cite{franken2013automatic} & Egyptian hieroglyph \\
Flying insects~\cite{kaggle_insect} & insect species \\
Arabic sign language~\cite{https://doi.org/10.48550/arxiv.2301.11932} & sign language \\
\bottomrule
\end{tabular}

\vspace{-4mm}
\end{wraptable}

%% file: ECCV_2026/sec/4_evaluation.tex
\section{Evaluation}
\label{sec:evaluation}
\subsection{Datasets}
\label{sec:eval_datasets}
We evaluate the proposed technique on a diverse collection of datasets spanning established few-shot benchmarks and novel specialized domains.

\paragraph{Standard benchmarks.}
We evaluate on standard few-shot classification benchmarks following \cite{liu2024, sav, fifty2023context}, including mini-ImageNet~\cite{vinyals2016matching}, a widely used general-purpose few-shot benchmark; CUB~\cite{wah2011caltech}, Aircraft~\cite{maji13fine-grained}, and Dogs~\cite{khosla2011novel}, which test fine-grained visual discrimination within natural categories; UCF101~\cite{soomro2012ucf101} (middle frame of video clip), representing action recognition under a video-to-image transfer. We further remove action images with strong object bias \cite{shvetsova2025unbiasing}. In DomainNet~\cite{Peng_2019_ICCV}, we sample episodes within a single domain at a time, excluding the \textit{real} domain to avoid overlap with the mini-ImageNet domain. 


\paragraph{Novel datasets.}
We introduce a benchmark suite of six visually distinctive datasets to probe model generalization under stronger domain shift. Namely, we consider the Lego bricks dataset (Lego) \cite{kaggle_lego}, industrial parts (Ind.) \cite{kaggle_industrial}, yoga poses (Yoga) \cite{kaggle_yoga}, ancient Egyptian hieroglyphs (Hiero.) \cite{franken2013automatic}, flying insects (Insect) \cite{kaggle_insect} and Arabic alphabet sign language (Sign) \cite{https://doi.org/10.48550/arxiv.2301.11932}. These cover highly specialized concepts as shown by the particularly low 0-shot recognition performance of pretrained MLLMs. We thus assume that those concepts are only sparsely represented in standard text-image pretraining, making them suitable test cases for evaluating the visual few-shot capabilities of pretrained MLLMs. 


\input{ECCV_2026/tab/SOTA_table}

\subsection{Experimental Details}

\paragraph{Models.}
We analyze three open-source state-of-the-art MLLMs of comparable scale (7--8B parameters): Qwen2.5-VL-7B-Instruct~\cite{bai2025qwen2}, Qwen3-VL-8B-Instruct~\cite{bai2025qwen3}, 
and InternVL3-8B~\cite{zhu2025internvl3}. All models are used in inference-only mode, with token-level 
scores
extracted directly from output logits as described in Eq.~\eqref{eq:pairwise}.

\paragraph{SFT Baseline.}
We use supervised fine-tuning (SFT) for Qwen3-VL-8B-Instruct with LoRA \cite{hu2022lora} on the semantic in-context few-shot formulation. The training data consists of 1000 5-way 1-shot episodes sampled from the miniImageNet base classes and the model is trained to output the correct query class. We fine-tune for 3 epochs with LoRA rank 16 and alpha 16, applying LoRA to all non-visual linear modules while keeping the visual modules frozen. Optimization uses a learning rate of ($5\times10^{-5}$), warmup ratio 0.03. The model is then evaluated directly on all target datasets without any further tuning. 

\paragraph{FSL Baselines.}
We compare against a diverse set of few-shot learning baselines. 
CLIP~\cite{lpclip} and SigLIP~\cite{zhai2023sigmoid} are vision-language models that support 0-shot classification via text-image cosine similarity, and few-shot classification via K-nearest neighbour (KNN) search over image embeddings. 
CLIP uses a ViT-B/32 backbone, while the SigLIP model variant is \texttt{base-patch16-224}. 
DINOv2 \texttt{base} model~\cite{oquab2023dinov2} and I-JEPA (ViT-H/14)~\cite{assran2023self} are vision-only self-supervised models, evaluated via KNN over frozen embeddings. 
Tip-Adapter~\cite{tipadapter} builds a training-free cache model from CLIP features (ViT-B/32).
CAML~\cite{fifty2023context} is a large-scale meta-learning method with a ViT-H encoder pre-trained on Laion-2B \cite{schuhmann2022laion} that performs in-context classification from frozen embeddings. 
ProKeR~\cite{proker} implements training-free kernel regression over pretrained CLIP features (ViT-B/32). 
Finally, SAVs~\cite{sav} directly leverage internal representations of MLLMs for training-free few-shot classification. We implement SAVs on top of Qwen2.5-VL-7B-Instruct, making it the closest baseline to our approach. 

\subsection{Comparison to State-of-the-art}
\label{sec:eval_sota}

Table~\ref{tab:sota} compares our method against few-shot baselines and recent MLLM-based approaches across standard and novel benchmarks, evaluated under both semantic and anonymized settings. 

\textbf{With semantic labels.}
On standard datasets, 0-shot MLLMs already achieve remarkably strong performance. For example, Qwen3-VL and InternVL3 reach above 90\% average accuracy without any support examples. Interestingly, 1-shot in-context inference often performs worse than 0-shot (e.g., InternVL3 drops from 90.7\% to 72.2\% average; Qwen2.5-VL from 92.2\% to 76.0\%). This indicates that adding support images does not guarantee effective adaptation and may even introduce interference when strong semantic priors dominate.

In contrast, on novel datasets, the trend reverses: 1-shot generally improves over 0-shot. For instance, Qwen3-VL increases from 57.0\% (0-shot avg) to 76.9\% (1-shot avg), suggesting that support examples become more beneficial as pretraining overlap decreases. This highlights the importance of evaluating beyond saturated standard benchmarks.

Across both dataset groups, our decompositional approach consistently outperforms 1-shot inference and matches or surpasses 0-shot performance on standard benchmarks, while substantially outperforming 0-shot performance on novel datasets. Adding dataset-specific domain information further improves results, yielding the best overall averages (e.g., 90.6\% for Qwen3-VL, 88.7\% for Qwen2.5-VL) and establishing new state-of-the-art few-shot performance. 

Compared to the FSL baseline methods, mainly Qwen3-VL models outperform current methods in the simple decomposed setting, while the other two models perform on par. This is expected for standard datasets, given the large-scale pretraining data, but the difference becomes more pronounced for novel datasets, where MLLMs significantly outperform our considered FSL baselines in most cases while remaining on par with the SFT baseline. 

\textbf{Without semantic labels (Anonymous).}
When semantic class names are removed, the performance of in-context 1-shot MLLMs drops dramatically. InternVL3 and Qwen3-VL collapse to near-random prediction (around 23–28\% total average in 5-way classification), revealing a heavy reliance on semantic priors rather than true support–query alignment. Qwen2.5-VL shows slightly more robustness but still degrades substantially.

In stark contrast, the decomposition approach remains highly effective in the anonymized regime. For all three MLLMs, our method restores strong few-shot performance. With Qwen3-VL, it surpasses the SFT baseline by 23.4 points. This shows that structured pairwise inference explicitly enforces visual correspondence and mitigates the semantic dominance.

Interestingly, also in this setting, where MLLMs cannot rely on their textual knowledge, they are able to outperform few-shot baseline methods. We again observe that, also here, the performance difference to the baseline models is smaller on the standard benchmarks, but widens for the novel benchmark setting. This shows that MLLMs might have learned general visual patterns that help them to generalize beyond the training distribution. 

Moreover, in Panel B of Table~\ref{tab:sota}, incorporating \textbf{domain information} leads to consistent overall gains across models and evaluation settings. The improvement is particularly clear on the novel datasets, suggesting that $D_{\text{info}}$ provides useful high-level context about the label space when the classes are less familiar. These results indicate that lightweight domain descriptions can complement decision-based prediction and improve recognition without additional training.

Overall, the results confirm that the proposed decomposed inference framework consistently enables robust few-shot adaptation across both semantic and anonymized settings, achieving state-of-the-art performance.



\subsection{Ablation}
\label{sec:eval_ablation}

\input{ECCV_2026/tab/dec_label_ablation_table}

We ablate three design choices that 
determine the effectiveness of our framework. 



\paragraph{Ablation of Decomposed Prompting.}
First, we analyze whether the performance gains stem from the binary decomposition itself or from the support–query comparison it introduces.
To this end, we consider a 0-shot decompose variant where no support image is given; the model answers: 
\texttt{“Is this image depicting $c_n$? Yes or No.”} 
In Table~\ref{tab:ablation}, 0-shot decomposition does not improve over standard 0-shot inference and can 
even perform slightly worse
(e.g., Qwen3-VL: 78.2\% → 76.8\%). 
This indicates that the gains of 1-shot decomposition do not arise from
the binary prompt structure, but from the support-query comparisons.

\paragraph{Ablation on Calibration Effects.}
Second, given that MLLMs exhibit strong language priors, we examine whether the performance of pairwise decomposition just stems from an implicit calibration. 
To this end, we consider an in-context 1-shot inference with calibration. Namely, we apply pointwise mutual information (PMI)-style calibration~\cite{zhao2021calibrate} to reduce option bias introduced by the prompt context.
For each candidate class, we compute the standard prediction given the query image $x^q$ and support set $\mathcal{S}$, and then subtract the score obtained when the query image is replaced by a placeholder text \texttt{“query image is omitted”} while keeping the rest of the prompt unchanged: $P(y|x^q,\mathcal{S})-P(y|\mathcal{S})$. 
As shown in Table~\ref{tab:ablation}, PMI-style calibration mitigates option bias in in-context inference and improves total accuracy (e.g., InternVL3: 62.6\% → 75.0\%; Qwen3-VL: 80.0\% → 86.2\%). 
However, it remains 
inferior to decomposition. Replacing N-way inference with pairwise decomposition consistently yields larger gains (e.g., Qwen3-VL: 86.2\% → 89.2\%), with 
particularly great improvements on novel datasets.




\paragraph{Evaluation of Logit scoring.}
\input{ECCV_2026/tab/logit_score_table}
Having established decomposition as the key component, we verify that its performance is robust to the choice of logit scoring rule used for ranking.  
Our pairwise inference ranks each support--query pair using the logit score of the \texttt{Yes} token (Eq.~\eqref{eq:pairwise}).
Table~\ref{tab:scoring_rules} compares alternative scoring strategies on Qwen2.5-VL, including: (1) using the model’s generated confidence score output for \texttt{Yes}; (2) minimizing the \texttt{No} logit score: Score(\texttt{No}); (3) a PMI-style correction Score(\texttt{Yes}) - Score(\texttt{No}); (4) the direct \texttt{Yes} logit score.

The exact prompt used to obtain the confidence score is provided in the supplementary material.
The three logit-based variants yield nearly identical results across all datasets, confirming 
that the method is robust to the choice of ranking rule.
In contrast, using the model's explicit confidence score leads to a substantial drop in accuracy (e.g., 73.3 on mini-ImageNet, 46.9 on Lego), indicating that generative confidence signals are less reliable for fine-grained pairwise matching. We provide further details on scoring in the supplement.

\subsection{5-way 5-shot Analysis}
\label{sec:eval_n_way_kshot}
\input{ECCV_2026/tab/5w5s_table}
We analyze the effect of increasing the number of shots under a fixed 5-way setting. 
The 5-way 5-shot episodes are constructed by extending the corresponding 5-way 1-shot episodes with four additional support samples per class, allowing for a controlled comparison (see also Equation~\ref{eq:pairwise}). Results are reported in Table~\ref{tab:5w_5s} for Yoga, Lego, and Industrial datasets.

While in-context learning benefits from a single support example, its performance deteriorates when moving to 5-shot, suggesting that longer in-context prompts introduce interference rather than improved adaptation.
In contrast, decomposition with $D_{\text{info}}$ consistently improves from 1-shot to 5-shot, achieving the best average accuracy (85.8\%), outperforming the SFT baseline. This behavior highlights a key advantage of decomposition: additional support images contribute positively through independent pairwise aggregation.


\subsection{N-way 1-shot Analysis}
\label{sec:eval_n_way}
\begin{figure}[t]
  \centering
  \begin{subfigure}[t]{0.49\linewidth}
    \centering
    \includegraphics[width=\linewidth]{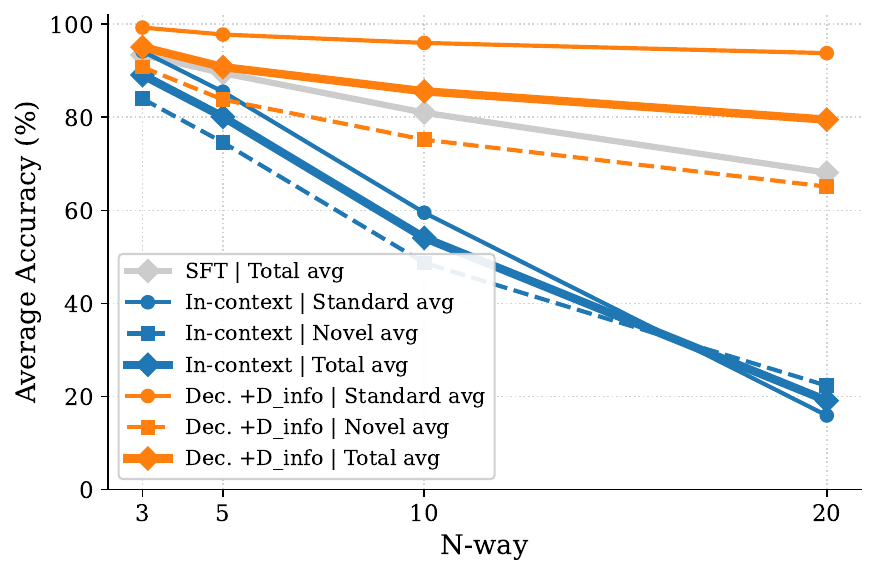}
    \caption{$N$-way-1-shot accuracy for comparing \textsc{In-context} and \textsc{Decompose} + domain info performance. We report the average accuracy for standard and novel datasets, as well as for all.}
    \label{fig:nway_scaling_acc}
  \end{subfigure}\hfill
  \begin{subfigure}[t]{0.49\linewidth}
    \centering
    \includegraphics[width=\linewidth]{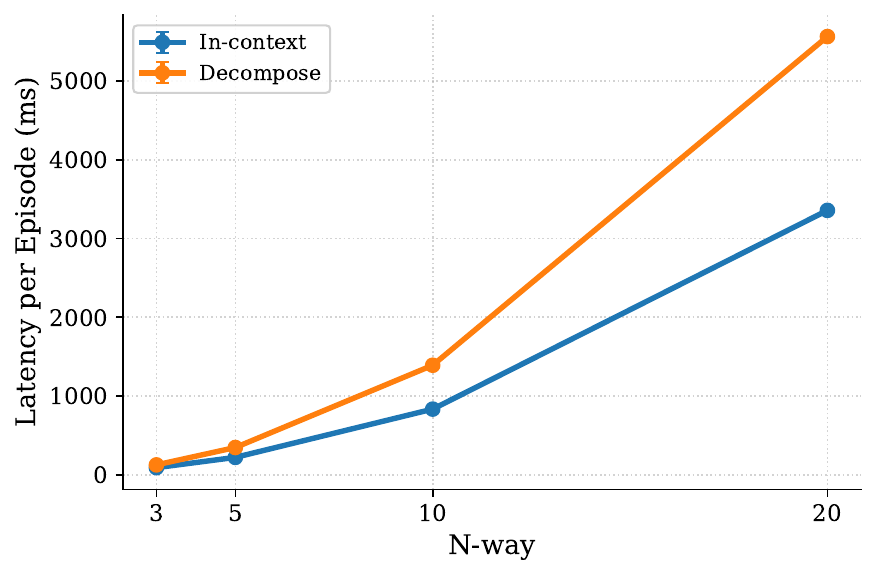}
    \caption{$N$-way-1-shot runtime analysis comparing \textsc{In-context} and \textsc{Decompose} + domain info. We report the mean episode latency for single-pass and decomposed batch inference (batch size 64).}
    \label{fig:nway_scaling_runtime}
  \end{subfigure}
  \caption{\textbf{Scaling with $N$-way using Qwen3-VL.} (a) Accuracy comparison between \textsc{In-context} (with and without SFT) and \textsc{Decompose} + domain info. (b) Corresponding runtime analysis under identical decoding and batching settings. $N \in \{3,5,10,20\}$.}
  \label{fig:nway_scaling}
\end{figure}

We further analyze the scalability with respect to the number of classes $N \in \{3,5,10,20\}$ under the 1-shot setting. 
In Fig.~\ref{fig:nway_scaling_acc}, In-context inference degrades sharply as $N$ increases. 
While performance is competitive at small $N$, accuracy drops substantially at 10- and 20-way episodes. 
In contrast, our decomposed method remains robust across all $N$, outperforming the SFT baseline. The performance gap widens as the task becomes more challenging, demonstrating that structured pairwise comparison scales more reliably than standard N-way prompting. 
This trend holds consistently across standard, novel, and averaged results, confirming that the improved scalability is not dataset-specific but inherent to the inference structure.

\subsection{Runtime}
\label{sec:eval_runtime}
The main limitation of decomposition is that it requires access to token logits and incurs $O(N)$ comparisons for an $N$-way problem. Thus, to investigate the computational cost difference, we measure runtime as end-to-end episode latency (ms) in the 1-shot $N$-way setting using Qwen3-VL ($N \in \{3,5,10,20\}$), averaged over 50 episodes per $N$. We compare in-context prompting, which requires one forward pass per episode, and with the proposed pairwise decomposition, which requires $N$ comparisons per episode. All experiments were conducted on a single NVIDIA H100 GPU (80GB) \cite{nvidia2022h100} using \texttt{Bfloat16} precision, deterministic decoding, left padding, CUDA synchronization, and a batch size of 64 for both episodes and generation. We report the mean latency per episode. 

Results in Figure \ref{fig:nway_scaling_runtime} show approximately linear scaling for both single inference and decomposition. The runtime gap is modest at small $N$ but widens at 10--20 way (approximately $1.7\times$ slower at 20-way), reflecting the computational trade-off introduced by structured decomposition. While those numbers reflect the single GPU setting, we argue that, especially for real-world scenarios, settings with larger $N$ or $K$ would most likely aim for a parallelization across multiple GPUs, as all forward paths can be processed independently, which would allow to mitigate longer runtime in those cases if needed.

%% file: ECCV_2026/tab/SOTA_table.tex
%
{
\hypersetup{hidelinks}

\providecolor{sotaStdBlue}{RGB}{210,225,245}
\providecolor{sotaNovelAmber}{RGB}{255,210,160}
\providecolor{sotaSecBlue}{RGB}{235,242,252}
\providecolor{sotaSecOrange}{RGB}{255,245,230}
\providecolor{sotaSFTText}{RGB}{130,130,130}
\providecolor{sotaDinfoTint}{RGB}{255,235,205}
\providecolor{almond}{RGB}{255,235,205}

\providecolor{sotaPanelGrey}{RGB}{235,231,223}  
\providecolor{sotaModelGrey}{RGB}{242,239,234}  
\providecolor{sotaModelTint}{RGB}{249,248,246}  

\providecommand{\sft}[1]{\textcolor{sotaSFTText}{#1}}
\providecommand{\panelrow}[1]{%
  \specialrule{0.8pt}{3pt}{1pt}
  \multicolumn{17}{l}{\cellcolor{sotaPanelGrey}\textbf{#1}}\\[-1pt]
  \specialrule{0.8pt}{1pt}{2pt}
}

\begin{table*}[!t]
\centering
\begingroup
\setlength{\hfuzz}{15pt}
\resizebox{\textwidth}{!}{%
\fontsize{5.6}{6.7}\selectfont
\setlength{\tabcolsep}{2.1pt}
\renewcommand{\arraystretch}{0.95}
\begin{tabular}{p{1.4cm}@{\hspace{2pt}}p{2.2cm} cccccc|c| cccccc|c|c}
\panelrow{Panel A: Main comparison}
\multicolumn{2}{l}{} &
  \multicolumn{7}{c}{\cellcolor{sotaStdBlue}\textbf{Standard Datasets}} &
  \multicolumn{7}{c}{\cellcolor{sotaNovelAmber}\textbf{Novel Datasets}} & \\[-2pt]
\cmidrule(lr){3-9}\cmidrule(lr){10-16}
\multicolumn{2}{l}{\textbf{Method}}
  & \textbf{mini} & \textbf{UCF} & \textbf{CUB} & \textbf{Air.} & \textbf{Dogs} & \textbf{Dom.} & \textbf{Avg}
  & \textbf{Lego} & \textbf{Ind.} & \textbf{Yoga} & \textbf{Hiero.} & \textbf{Insect} & \textbf{Sign}
  & \textbf{Avg} & \textbf{Total} \\

\midrule

\multicolumn{17}{l}{\cellcolor{sotaModelGrey}\textbf{With labels (Semantic)}} \\[-1pt]
\midrule

\multicolumn{17}{l}{\cellcolor{sotaModelTint}\textit{Reported results}} \\
SAVs \cite{sav}               & 1-shot & --   & --   & 98.7 & --   & --   & --   & --   & --   & --   & --   & --   & --   & --   & --   & --  \\
CAML \cite{fifty2023context} & 1-shot & 96.2 & --   & 91.8 & 63.3 & --   & --   & --   & --   & --   & --   & --   & --   & --   & --   & --  \\
GFSL \cite{liu2024}          & 1-shot & 98.2 & --   & 96.6 & 96.6 & 96.7 & --   & --   & --   & --   & --   & --   & --   & --   & --   & --  \\[2pt]

\multicolumn{17}{l}{\cellcolor{sotaModelTint}\textit{Reproduced baselines}} \\
        CLIP \cite{lpclip}                       & 0-shot      & 95.5 & 81.2 & 94.0 & 72.9 & 90.4 & 86.2 & 86.7 & 30.3 & 31.7 & 31.3 & 39.9 & 57.7 & 20.0 & 35.2 & 60.9 \\
        ProKeR \cite{proker}                     & 1-shot      & 98.1 & 89.0 & 94.7 & 73.4 & 92.5 & \textbf{90.3} & 89.3 & 49.7 & 51.7 & 43.2 & 68.1 & 70.3 & 38.2 & 53.5 & 71.4 \\
        SigLip \cite{zhai2023sigmoid}            & 0-shot      & 98.3 & 85.4 & \textbf{97.5} & 79.0 & \textbf{98.7} & 88.2 & 91.5 & 63.0 & 49.4 & \textbf{91.8} & 47.2 & 72.9 & 20.2 & 57.4 & 74.5 \\
        SAVs \cite{sav}                           & 1-shot      & 98.4 & \textbf{94.4} & 96.0 & \textbf{89.4} & 93.1 & 86.2 & \textbf{92.9} & \textbf{68.2} & 57.8 & 62.4 & 64.4 & 80.4 & 28.8 & 60.3 & 76.6 \\
        CAML \cite{fifty2023context}             & 1-shot      & \textbf{98.7} & 90.7 & 95.3 & 86.9 & 94.9 & 81.2 & 91.3 & 62.3 & \textbf{78.0} & 25.2 & \textbf{84.4} & \textbf{95.4} & \textbf{52.3} & \textbf{66.3} & \textbf{78.8} \\[1pt]

\multicolumn{17}{l}{\cellcolor{sotaModelTint}\textit{InternVL3}} \\
                                                 & 0-shot      & \textbf{99.1} & \textbf{95.7} & 89.6 & 73.8 & \textbf{93.7} & \textbf{92.0} & 90.7 & 53.1 & 60.8 & 58.5 & 45.2 & 68.2 & 18.8 & 50.8 & 70.7 \\
                                                 & 1-shot      & 76.2 & 94.8 & 83.2 & 30.3 & 86.6 & 61.8 & 72.2 & 49.0 & 64.9 & 45.2 & 35.6 & 82.4 & 20.0 & 49.5 & 60.8 \\
\rowcolor{almond}                                & 1-shot dec. & 98.4 & 94.5 & \textbf{94.1} & \textbf{75.7} & 92.7 & 89.8 & \textbf{90.9} & \textbf{62.8} & \textbf{75.0} & \textbf{73.6} & \textbf{75.7} & \textbf{89.0} & \textbf{58.7} & \textbf{72.5} & \textbf{81.7} \\[1pt]

\multicolumn{17}{l}{\cellcolor{sotaModelTint}\textit{Qwen2.5-VL}} \\
                                                 & 0-shot      & 97.2 & 93.0 & 96.4 & 89.5 & 91.8 & 85.3 & 92.2 & 50.1 & 42.9 & 50.7 & 43.1 & 80.0 & 20.7 & 47.9 & 70.1 \\
                                                 & 1-shot      & 88.1 & 84.7 & 66.6 & 61.9 & 81.0 & 73.6 & 76.0 & 54.9 & 49.9 & 67.9 & 66.0 & 88.8 & 59.3 & 64.5 & 70.2 \\
\rowcolor{almond}                                & 1-shot dec. & \textbf{98.5} & \textbf{93.7} & \textbf{96.9} & \textbf{94.6} & \textbf{95.8} & \textbf{90.5} & \textbf{95.0} & \textbf{65.5} & \textbf{73.5} & \textbf{80.6} & \textbf{78.1} & \textbf{93.3} & \textbf{67.7} & \textbf{76.5} & \textbf{85.7} \\[1pt]

\multicolumn{17}{l}{\cellcolor{sotaModelTint}\textit{Qwen3-VL}} \\
                                                 & 0-shot      & \textbf{99.1} & 96.6 & 96.5 & 89.1 & \textbf{96.4} & \textbf{92.6} & 95.1 & 63.4 & 62.8 & 66.0 & 48.1 & 80.1 & 21.3 & 57.0 & 76.0 \\
                                                 & 1-shot      & 85.3 & 92.7 & 85.0 & 75.3 & 86.2 & 84.6 & 84.9 & 66.9 & 80.3 & 74.5 & 82.4 & \textbf{89.1} & 68.4 & 76.9 & 80.9 \\
\rowcolor{almond}                                & 1-shot dec. & 98.9 & \textbf{97.3} & \textbf{96.8} & \textbf{92.2} & 96.1 & 91.9 & \textbf{95.5} & \textbf{70.6} & \textbf{81.4} & \textbf{83.4} & \textbf{89.5} & \textbf{89.1} & \textbf{70.6} & \textbf{79.1} & \textbf{87.3} \\
                                                 & \sft{1-shot SFT} & \sft{98.9} & \sft{97.6} & \sft{97.6} & \sft{90.1} & \sft{96.9} & \sft{92.9} & \sft{95.7} & \sft{77.9} & \sft{77.9} & \sft{81.7} & \sft{81.6} & \sft{95.5} & \sft{70.1} & \sft{80.8} & \sft{88.2} \\

\midrule

\multicolumn{17}{l}{\cellcolor{sotaModelGrey}\textbf{Without labels (Anonymous)}} \\[-1pt]
\midrule

\multicolumn{17}{l}{\cellcolor{sotaModelTint}\textit{Reproduced baselines}} \\
        I-JEPA \cite{assran2023self}             & 1-shot      & 76.8 & 75.5 & 53.8 & 32.5 & 79.1 & 39.8 & 59.6 & 45.1 & 42.1 & 60.0 & 68.4 & 45.1 & 31.3 & 48.7 & 54.1 \\
        Tip-A. \cite{tipadapter}                 & 1-shot      & \textbf{97.7} & 82.2 & 93.9 & 70.9 & 90.2 & \textbf{87.0} & \textbf{87.0} & 31.2 & 39.5 & 33.2 & 42.2 & 62.1 & 21.5 & 38.3 & 62.6 \\
        CLIP \cite{lpclip}                       & 1-shot      & 86.5 & 83.0 & 80.6 & 61.0 & 66.7 & 63.9 & 73.6 & 51.3 & 51.5 & 35.3 & 80.4 & 76.8 & 41.5 & 56.1 & 64.9 \\
        DINO \cite{oquab2023dinov2}              & 1-shot      & 93.0 & \textbf{95.2} & \textbf{97.8} & 64.6 & \textbf{95.5} & 74.1 & 86.7 & 65.0 & 73.6 & 52.1 & 85.1 & 66.0 & 40.2 & 63.7 & 75.2 \\
        SigLip \cite{zhai2023sigmoid}            & 1-shot      & 91.3 & 88.6 & 90.4 & \textbf{85.2} & 84.7 & 70.1 & 85.1 & \textbf{70.3} & \textbf{75.7} & \textbf{62.4} & \textbf{87.9} & \textbf{91.0} & \textbf{48.2} & \textbf{72.6} & \textbf{78.8} \\[1pt]

\multicolumn{17}{l}{\cellcolor{sotaModelTint}\textit{InternVL3}} \\
                                                 & 1-shot      & 22.9 & 36.0 & 20.4 & 21.0 & 21.0 & 19.9 & 23.5 & 23.0 & 27.5 & 23.4 & 21.0 & 23.3 & 21.9 & 23.4 & 23.4 \\
\rowcolor{almond}                                & 1-shot dec. & \textbf{94.8} & \textbf{92.8} & \textbf{90.6} & \textbf{70.8} & \textbf{91.4} & \textbf{78.8} & \textbf{86.5} & \textbf{55.4} & \textbf{81.2} & \textbf{51.9} & \textbf{92.1} & \textbf{86.0} & \textbf{76.2} & \textbf{73.8} & \textbf{80.2} \\[1pt]

\multicolumn{17}{l}{\cellcolor{sotaModelTint}\textit{Qwen2.5-VL}} \\
                                                 & 1-shot      & 23.0 & 70.4 & 38.4 & 47.9 & 43.7 & 30.9 & 42.4 & 41.3 & 49.4 & 57.2 & 57.1 & \textbf{88.8} & 68.8 & 60.4 & 51.4 \\
\rowcolor{almond}                                & 1-shot dec. & \textbf{97.0} & \textbf{94.9} & \textbf{96.2} & \textbf{92.8} & \textbf{92.7} & \textbf{82.9} & \textbf{92.8} & \textbf{60.8} & \textbf{73.9} & \textbf{74.6} & \textbf{86.2} & 72.1 & \textbf{76.5} & \textbf{74.0} & \textbf{83.4} \\[1pt]

\multicolumn{17}{l}{\cellcolor{sotaModelTint}\textit{Qwen3-VL}} \\
                                                 & 1-shot      & 21.4 & 26.6 & 38.0 & 39.4 & 27.7 & 20.5 & 28.9 & 25.6 & 31.1 & 21.3 & 30.0 & 23.6 & 28.1 & 26.6 & 27.8 \\
\rowcolor{almond}                                & 1-shot dec. & \textbf{94.9} & \textbf{93.0} & \textbf{87.0} & \textbf{91.1} & \textbf{88.1} & \textbf{79.1} & \textbf{88.9} & \textbf{63.7} & \textbf{81.2} & \textbf{67.5} & \textbf{94.5} & \textbf{33.9} & \textbf{72.6} & \textbf{68.9} & \textbf{78.9} \\
                                                 & \sft{1-shot SFT} & \sft{51.4} & \sft{51.8} & \sft{80.3} & \sft{65.1} & \sft{62.3} & \sft{26.8} & \sft{56.3} & \sft{44.2} & \sft{51.1} & \sft{51.4} & \sft{58.4} & \sft{68.2} & \sft{54.4} & \sft{54.6} & \sft{55.5} \\

\midrule
\panelrow{Panel B: With dataset-specific domain information}

\multicolumn{2}{l}{\textbf{Method}}
  & \textbf{mini} & \textbf{UCF} & \textbf{CUB} & \textbf{Air.} & \textbf{Dogs} & \textbf{Dom.} & \textbf{Avg}
  & \textbf{Lego} & \textbf{Ind.} & \textbf{Yoga} & \textbf{Hiero.} & \textbf{Insect} & \textbf{Sign}
  & \textbf{Avg} & \textbf{Total} \\
\midrule

\multicolumn{17}{l}{\cellcolor{sotaModelGrey}\textbf{With labels (Semantic)}} \\[-1pt]
\midrule

\multicolumn{17}{l}{\cellcolor{sotaModelTint}\textit{InternVL3}} \\
\rowcolor{almond}        & 1-shot dec.              & \textbf{98.4} & 94.5 & 94.1 & 75.7 & 92.7 & \textbf{89.8} & 90.9 & 62.8 & 75.0 & 73.6 & 75.7 & 89.0 & 58.7 & 72.5 & 81.7 \\
\rowcolor{almond} & 1-shot dec. +$D_{\text{info}}$       & 98.0 & \textbf{94.9} & \textbf{95.1} & \textbf{76.6} & \textbf{93.0} & \textbf{89.8} & \textbf{91.2} & \textbf{64.7} & \textbf{75.3} & \textbf{80.1} & \textbf{77.3} & \textbf{91.8} & \textbf{65.1} & \textbf{75.7} & \textbf{83.5} \\[2pt]

\multicolumn{17}{l}{\cellcolor{sotaModelTint}\textit{Qwen2.5-VL}} \\
\rowcolor{almond}        & 1-shot dec.              & \textbf{98.5} & 93.7 & 96.9 & 94.6 & 95.8 & 90.5 & 95.0 & 65.5 & 73.5 & 80.6 & 78.1 & 93.3 & 67.7 & 76.5 & 85.7 \\
\rowcolor{almond} & 1-shot dec. +$D_{\text{info}}$       & \textbf{98.5} & \textbf{94.3} & \textbf{97.6} & \textbf{95.8} & \textbf{96.5} & \textbf{91.0} & \textbf{95.6} & \textbf{66.4} & \textbf{74.2} & \textbf{84.3} & \textbf{88.1} & \textbf{97.0} & \textbf{80.9} & \textbf{81.8} & \textbf{88.7} \\[2pt]

\multicolumn{17}{l}{\cellcolor{sotaModelTint}\textit{Qwen3-VL}} \\
\rowcolor{almond}        & 1-shot dec.              & 98.9 & \textbf{97.3} & 96.8 & 92.2 & 96.1 & 91.9 & 95.5 & 70.6 & \textbf{81.4} & 83.4 & 89.5 & 89.1 & 70.6 & 79.1 & 87.3 \\
\rowcolor{almond} & 1-shot dec. +$D_{\text{info}}$       & \textbf{99.0} & \textbf{97.3} & \textbf{97.9} & \textbf{93.6} & \textbf{96.6} & \textbf{92.0} & \textbf{96.1} & \textbf{72.9} & 80.5 & \textbf{88.3} & \textbf{90.2} & \textbf{97.4} & \textbf{82.0} & \textbf{85.2} & \textbf{90.6} \\[2pt]

\midrule
\multicolumn{17}{l}{\cellcolor{sotaModelGrey}\textbf{Without labels (Anonymous)}} \\[-1pt]
\midrule

\multicolumn{17}{l}{\cellcolor{sotaModelTint}\textit{InternVL3}} \\
\rowcolor{almond}        & 1-shot dec.              & 94.8 & 92.8 & 90.6 & 70.8 & \textbf{91.4} & 78.8 & 86.5 & 55.4 & \textbf{81.2} & 51.9 & \textbf{92.1} & 86.0 & 76.2 & 73.8 & 80.2 \\
\rowcolor{almond} & 1-shot dec. +$D_{\text{info}}$       & \textbf{95.1} & \textbf{95.0} & \textbf{95.3} & \textbf{77.0} & 90.7 & \textbf{80.6} & \textbf{89.0} & \textbf{61.2} & 78.2 & \textbf{84.8} & 90.3 & \textbf{93.8} & \textbf{86.9} & \textbf{82.5} & \textbf{85.7} \\[2pt]

\multicolumn{17}{l}{\cellcolor{sotaModelTint}\textit{Qwen2.5-VL}} \\
\rowcolor{almond}        & 1-shot dec.              & 97.0 & \textbf{94.9} & 96.2 & 92.8 & 92.7 & 82.9 & 92.8 & 60.8 & \textbf{73.9} & 74.6 & \textbf{86.2} & 72.1 & 76.5 & 74.0 & 83.4 \\
\rowcolor{almond} & 1-shot dec. +$D_{\text{info}}$       & \textbf{97.2} & 92.4 & \textbf{97.4} & \textbf{95.2} & \textbf{94.7} & \textbf{84.3} & \textbf{93.5} & \textbf{63.0} & 72.1 & \textbf{84.3} & 83.0 & \textbf{97.5} & \textbf{82.0} & \textbf{80.3} & \textbf{86.9} \\[2pt]

\multicolumn{17}{l}{\cellcolor{sotaModelTint}\textit{Qwen3-VL}} \\
\rowcolor{almond}        & 1-shot dec.              & 94.9 & 93.0 & 87.0 & 91.1 & 88.1 & \textbf{79.1} & 88.9 & 63.7 & \textbf{81.2} & 67.5 & \textbf{94.5} & 33.9 & 72.6 & 68.9 & 78.9 \\
\rowcolor{almond} & 1-shot dec. +$D_{\text{info}}$       & \textbf{95.2} & \textbf{96.6} & \textbf{96.8} & \textbf{92.4} & \textbf{95.4} & 78.7 & \textbf{92.5} & \textbf{68.9} & 80.3 & \textbf{89.6} & 91.2 & \textbf{98.2} & \textbf{86.6} & \textbf{85.8} & \textbf{89.2} \\

\bottomrule
\end{tabular}
}
\endgroup
\vspace{2mm}
\caption{Comparison with state-of-the-art on standard/novel benchmarks for 5-way 1-shot, resp. 0-shot if indicated. Panel A reports the main comparison; Panel B reports the effect of dataset-specific domain information $D_{\text{info}}$. \textit{dec.}~=~decompose and \textit{SFT}~=~supervised fine-tuning. Best results are highlighted \textbf{bold} per subsection. Colored rows mark our method variants.}
\label{tab:sota}
\end{table*}
}

%% file: ECCV_2026/tab/dec_label_ablation_table.tex
%

\providecolor{sotaStdBlue}{RGB}{210,225,245}
\providecolor{sotaNovelAmber}{RGB}{255, 210, 160}
\providecolor{sotaModelTint}{RGB}{248,248,248}
\providecolor{sotaModelGrey}{RGB}{232,232,232}

\begin{table}[t]
\centering
\begingroup\setlength{\hfuzz}{15pt}\resizebox{\textwidth}{!}{%
\fontsize{7}{9}\selectfont
\setlength{\tabcolsep}{3.5pt}
\renewcommand{\arraystretch}{0.97}
\setlength{\hfuzz}{15pt}
\begin{tabular}{p{3.3cm} ccc|c|c ccc|c|c c}
\toprule
\multicolumn{1}{l}{} &
  \multicolumn{4}{c}{\cellcolor{sotaStdBlue}\scalebox{0.8}{\textbf{Standard Datasets}}} & &
  \multicolumn{4}{c}{\cellcolor{sotaNovelAmber}\scalebox{0.8}{\textbf{Novel Datasets}}} & \\[-2pt]
\cmidrule(lr){2-5}\cmidrule(lr){7-10}
\multicolumn{1}{l}{\textbf{Method}}
  & \textbf{mini} & \textbf{CUB} & \textbf{Dogs} & \textbf{Avg}
  & & \textbf{Lego} & \textbf{Yoga} & \textbf{Hiero.} & \textbf{Avg}
  & \textbf{Total} \\
\midrule

\multicolumn{11}{l}{\cellcolor{sotaModelGrey}\textbf{With labels (Semantic)}} \\[-1pt]
\midrule

\multicolumn{11}{l}{\cellcolor{sotaModelTint}\textit{InternVL3}} \\
\quad 0-shot                 & \textbf{99.1} & 89.6 & \textbf{93.7} & 94.1 &  & 53.1 & 58.5 & 45.2 & 52.3 & 73.2 \\
\quad 0-shot dec.            & 97.0 & 88.2 & 91.5 & 92.2 &  & 49.4 & 53.9 & 42.4 & 48.6 & 70.4 \\
\quad 1-shot \textcolor{gray}{(baseline)} & 76.2 & 83.2 & 86.6 & 82.0 &  & 49.0 & 45.2 & 35.6 & 43.3 & 62.6 \\
\quad 1-shot calibrated      & 94.0 & \textbf{95.7} & 90.9 & 93.5 &  & \textbf{65.4} & 56.6 & 47.2 & 56.4 & 75.0 \\
\rowcolor{almond}\quad 1-shot dec. \textcolor{gray}{(ours)}           & \underline{98.4} & 94.1 & 92.7 & \underline{95.1} &  & 62.8 & \underline{73.6} & \underline{75.7} & \underline{70.7} & \underline{82.9} \\
\rowcolor{almond}\quad 1-shot dec.+$D_{\text{info}}$ \textcolor{gray}{(ours)}     & 98.0 & \underline{95.1} & \underline{93.0} & \textbf{95.4} &  & \underline{64.7} & \textbf{80.1} & \textbf{77.3} & \textbf{74.0} & \textbf{84.7} \\[2pt]
\multicolumn{11}{l}{\cellcolor{sotaModelTint}\textit{Qwen2.5-VL}} \\
\quad 0-shot                 & 97.2 & 96.4 & 91.8 & 95.1 &  & 50.1 & 50.7 & 43.1 & 48.0 & 71.5 \\
\quad 0-shot dec.            & \underline{97.6} & 96.4 & \underline{96.0} & 96.7 &  & \underline{65.5} & 56.1 & 45.7 & 55.8 & 76.2 \\
\quad 1-shot \textcolor{gray}{(baseline)} & 88.1 & 66.6 & 81.0 & 78.6 &  & 54.9 & 67.9 & 66.0 & 62.9 & 70.8 \\
\quad 1-shot calibrated      & 94.8 & 96.1 & 92.7 & 94.5 &  & 64.1 & 70.7 & 72.7 & 69.2 & 81.8 \\
\rowcolor{almond}\quad 1-shot dec. \textcolor{gray}{(ours)}            & \textbf{98.5} & \underline{96.9} & 95.8 & \underline{97.1} &  & \underline{65.5} & \underline{80.6} & \underline{78.1} & \underline{74.7} & \underline{85.9} \\
\rowcolor{almond}\quad 1-shot dec.+$D_{\text{info}}$ \textcolor{gray}{(ours)}        & \textbf{98.5} & \textbf{97.6} & \textbf{96.5} & \textbf{97.5} &  & \textbf{66.4} & \textbf{84.3} & \textbf{88.1} & \textbf{79.6} & \textbf{88.5} \\[2pt]
\multicolumn{11}{l}{\cellcolor{sotaModelTint}\textit{Qwen3-VL}} \\
\quad 0-shot                 & \textbf{99.1} & 96.5 & \underline{96.4} & \underline{97.3} &  & 63.4 & 66.0 & 48.1 & 59.2 & 78.2 \\
\quad 0-shot dec.            & 96.2 & 96.0 & 95.9 & 96.0 &  & 63.4 & 62.7 & 46.7 & 57.6 & 76.8 \\
\quad 1-shot \textcolor{gray}{(baseline)} & 85.3 & 85.0 & 86.2 & 85.5 &  & 66.9 & 74.5 & 82.4 & 74.6 & 80.0 \\
\quad 1-shot calibrated      & 94.5 & 93.2 & 92.4 & 93.4 &  & \underline{70.9} & 80.0 & 85.8 & 78.9 & 86.2 \\
\rowcolor{almond}\quad 1-shot dec. \textcolor{gray}{(ours)}            & 98.9 & \underline{96.8} & 96.1 & \underline{97.3} &  & 70.6 & \underline{83.4} & \underline{89.5} & \underline{81.2} & \underline{89.2} \\
\rowcolor{almond}\quad 1-shot dec.+$D_{\text{info}}$ \textcolor{gray}{(ours)}        & \underline{99.0} & \textbf{97.9} & \textbf{96.6} & \textbf{97.8} &  & \textbf{72.9} & \textbf{88.3} & \textbf{90.2} & \textbf{83.8} & \textbf{90.8} \\

\bottomrule
\end{tabular}
}\endgroup
\vspace{2mm}
\caption{Ablation on decomposition and context across standard and novel benchmarks (5-way). Top-1 accuracy (\%). \textit{dec.}~=~decompose, \textit{$D_{\text{info}}$}~=~dataset specific domain info. Best and second-best results are highlighted \textbf{bold} and \underline{underlined} per section.}
\label{tab:ablation}
\end{table}

%% file: ECCV_2026/tab/logit_score_table.tex
\begin{table}[t]
\centering
\setlength{\tabcolsep}{4pt}
\small
\resizebox{\textwidth}{!}{%
\begin{tabular}{l ccc|c| ccc|c|c}
\toprule
\multicolumn{1}{l}{} &
  \multicolumn{4}{c}{\cellcolor{sotaStdBlue}\scalebox{0.7}{\textbf{Standard Datasets}}} &
  \multicolumn{4}{c}{\cellcolor{sotaNovelAmber}\scalebox{0.7}{\textbf{Novel Datasets}}} & \\[-2pt]
\cmidrule(lr){2-5}\cmidrule(lr){6-9}
Scoring rule & mini & CUB & Dogs & Avg & Lego & Yoga & Hiero. & Avg & Total \\
\midrule
(1) $\arg\max$[Confidence(\texttt{Yes})]                     & 73.3 & 84.6 & 86.6 & 81.5 & 46.9 & 71.2 & 68.2 & 62.1 & 71.8 \\
(2) $\arg\min$[Score(\texttt{No})]                           & 97.0 & 96.1 & 92.9 & 95.3 & 64.1 & 74.4 & 85.7 & 74.7 & 85.0 \\
(3) $\arg\max$[Score(\texttt{Yes}) - Score(\texttt{No})] & 97.1 & 96.2 & 92.6 & 95.3 & 63.9 & 74.4 & 86.1 & 74.8 & 85.1 \\
(4) $\arg\max$[Score(\texttt{Yes})]                             & 97.1 & 96.1 & 92.7 & 95.3 & 64.1 & 74.6 & 86.2 & 75.0 & 85.1 \\
\bottomrule
\end{tabular}
}
\vspace{2mm}
\caption{Comparison of pairwise scoring rules for Qwen2.5-VL in the 5-way 1-shot decomposed setting without semantic labels. All logit-style scoring variants perform nearly identically, confirming the scoring choice (4) of our method.}
\label{tab:scoring_rules}
\vspace{-5mm}
\end{table}

%% file: ECCV_2026/tab/5w5s_table.tex
\providecolor{sotaSFTText}{RGB}{130,130,130}
\providecommand{\sft}[1]{\textcolor{sotaSFTText}{#1}}

\begin{table}[t]
\centering
\setlength{\tabcolsep}{4pt}
\small
\begin{tabular}{l cc cc cc |cc}
\toprule
\multicolumn{1}{l}{} &
\multicolumn{8}{c}{\cellcolor{sotaNovelAmber}\scalebox{0.7}{\textbf{Novel Datasets}}} \\[-2pt]
\cmidrule(lr){2-9}
\multirow{2}{*}{Method}
  & \multicolumn{2}{c}{Yoga}
  & \multicolumn{2}{c}{Lego}
  & \multicolumn{2}{c}{Industrial}
  & \multicolumn{2}{c}{Avg} \\
\cmidrule(lr){2-3} \cmidrule(lr){4-5} \cmidrule(lr){6-7} \cmidrule(lr){8-9}
  & 5w-1s & 5w-5s & 5w-1s & 5w-5s & 5w-1s & 5w-5s & 5w-1s & 5w-5s \\
\midrule
In-context                    & 74.5 & 45.9 & 66.9 & 43.9 &  80.3 & 60.9 & 73.9 & 50.2 \\
Decompose + $D_{\text{info}}$        & \textbf{88.3} &  \textbf{91.8} &  \textbf{72.9} &  \textbf{80.7} &  \textbf{80.5} & \textbf{85.0} &  \textbf{80.6} &  \textbf{85.8} \\
\sft{SFT baseline}                  & \sft{81.5} & \sft{88.0} & \sft{77.6} & \sft{77.3} &  \sft{78.2} & \sft{82.4} & \sft{79.1} & \sft{82.6} \\
\bottomrule
\end{tabular}
\vspace{2mm}
\caption{Few-shot classification accuracy (\%) on Yoga, Lego, and Industrial datasets under semantic 5-way 1-shot and 5-way 5-shot settings using Qwen3-VL.}
\label{tab:5w_5s}
\end{table}

%% file: ECCV_2026/sec/5_conclusion.tex
\section{Conclusion}
\label{sec:Conclusion}

We presented a simple yet effective framework for few-shot image classification with multimodal large language models based on \textit{Decompose, Compare, Decide} (DeCoDe). Instead of prompting the model to directly select among class names, our method reformulates classification as binary support--query matching, using the logit of the \texttt{Yes} response as a similarity score aggregated across support examples. This structured inference encourages direct visual comparison and reduces reliance on semantic priors from class labels.

Across both standard and novel benchmarks, the proposed decomposition consistently improves performance over conventional in-context prompting and competitive training-free baselines, achieving state-of-the-art results. The gains are particularly pronounced in the anonymized setting where semantic labels are removed, revealing that structured pairwise comparison more effectively leverages the support set.

These findings suggest a practical inference-time adaptation strategy: users can define new visual categories using a few support images, while the MLLM classifies queries by explicitly comparing them against the provided references without parameter updates.

Beyond the inference method, we introduce a controlled evaluation protocol that disentangles semantic priors from visual adaptation by comparing zero-shot and few-shot regimes with and without semantic labels across diverse datasets. Our results show that standard in-context prompting relies heavily on semantic cues, while the proposed decomposition better captures support adaptation.



\newpage

%% file: ECCV_2026/sec/6_supplement.tex
\section{Supplementary Materials}
\label{sec:supplementary}


\subsection{Dataset Details}

\input{ECCV_2026/tab/dataset_details}

\FloatBarrier

\begin{figure}[!t]
\centering
\includegraphics[width=0.9\linewidth]{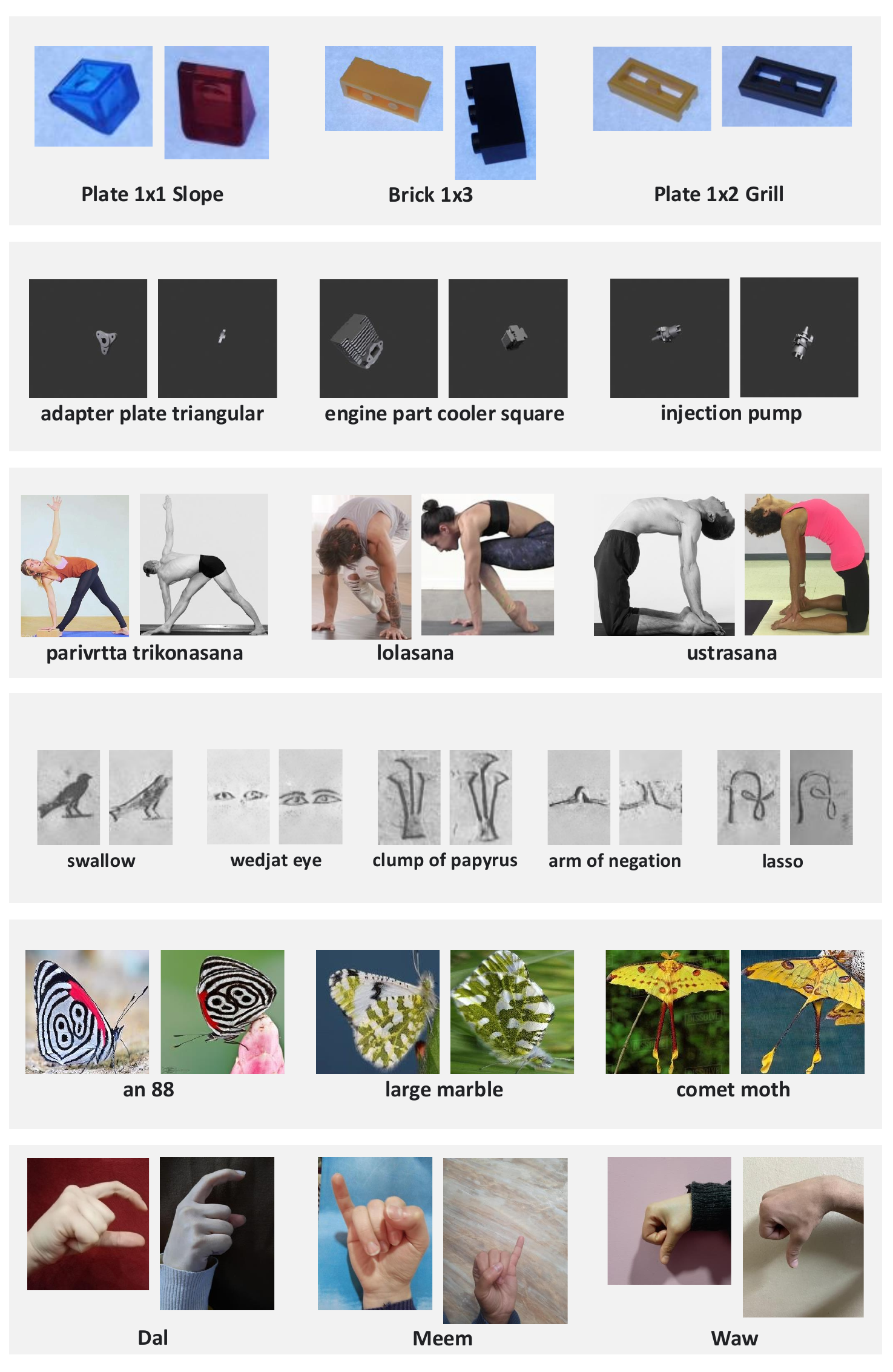}
\caption{
Example images from the novel datasets (ordered top-down): Lego bricks \cite{kaggle_lego}, Industrial parts \cite{kaggle_industrial}, Yoga \cite{kaggle_yoga}, Egyptian hieroglyphs \cite{franken2013automatic}, Flying insects \cite{kaggle_insect}, Arabic sign language \cite{https://doi.org/10.48550/arxiv.2301.11932}. 
}
\label{fig:example_images}
\end{figure}

Table \ref{tab:dataset_details} provides an overview of the datasets used for few-shot evaluation in this work. In the Dogs~\cite{khosla2011novel} dataset, to reduce background noise, we crop the dog image using the official bounding box annotation. In the Aircraft~\cite{maji13fine-grained} dataset, we use the granularity of variant as class, forming a total of 100 classes. In the Arabic sign language \cite{https://doi.org/10.48550/arxiv.2301.11932} dataset, we resize the images to have a width of 340 without distortion for lower computational costs. The Industrial parts \cite{kaggle_industrial} dataset contains 10 types of industrial products. Each is represented as a CAD object and rendered from various angles. We provide example images from the novel datasets in Figure \ref{fig:example_images}.

\subsection{Implementation Details}
\label{subsec:implementation_details}

\paragraph{Foundational Model Implementation.}
We compare against four foundational vision baselines: SigLIP (\texttt{base-patch16-224})~\cite{zhai2023sigmoid} and CLIP (\texttt{ViT-B/32})~\cite{lpclip}, both vision-language models with dual encoders that support 0-shot and 1-shot evaluation; and two vision-only models, DINOv2 (\texttt{base})~\cite{oquab2023dinov2} and I-JEPA (\texttt{ViT-H/14})~\cite{assran2023self} , which support only 1-shot. For 0-shot, CLIP and SigLIP use text prompts of the form \texttt{"This is a photo of \{class\}."} and predict via argmax over image--text similarity (softmax for CLIP, sigmoid for SigLIP). For 1-shot, all models extract L2-normalized image embeddings: CLS token for DINOv2, pooler or patch-mean for I-JEPA, and \texttt{get\_image\_features} for CLIP and SigLIP. Then assign each query to its nearest support-set prototype in cosine similarity. All models run in float16 without fine-tuning. 

\paragraph{CAML Implementation.}
CAML~\cite{fifty2023context} is an in-context meta-learner for 1-shot classification without fine-tuning.
We use CAML-Laion2b (frozen ViT-H/14, Laion-2B pretrained).
Support and query images are encoded by the frozen backbone; features are arranged as $[\mathbf{q}, \mathbf{s}_1, \ldots, \mathbf{s}_{N}]$ and each position is concatenated with a label embedding (query: unknown embedding; support: class embeddings).
The transformer encoder outputs logits over $N$ classes for the query, and the prediction is obtained by $\arg\max$ over the logits.

\paragraph{ProKeR Implementation.}
ProKeR~\cite{proker} is a training-free few-shot method that corrects CLIP’s zero-shot logits using the support set.
We use frozen CLIP (ViT-B/32) with prompts \texttt{"a photo of a \{class\}."}
On the support set, ProKeR fits the gap between ground-truth labels and zero-shot predictions using an RBF kernel over normalized image features, then interpolates this correction to the query images.
Hyperparameters: $\beta = 1.0$, $\lambda = 0.1$ (default).

\paragraph{SAVs Implementation.}
SAVs~\cite{sav} uses sparse attention activations from a frozen LMM (Qwen2.5-VL-7B) for few-shot classification.
Each sample is a single image and a text prompt (e.g., \texttt{"What class is shown? Answer with the class name only."}); there are no in-prompt few-shot examples.
We extract attention output vectors at the last input token, average them per class over the support set to form class prototypes, and select the top 20 heads that best predict support labels.
For each query, each head votes for the class with the highest cosine similarity to that head's prototype; the majority vote is the final prediction.

\paragraph{Tip-Adapter Implementation.}
Tip-Adapter~\cite{tipadapter} is a training-free few-shot method that augments zero-shot CLIP with a support-set cache.
We use frozen CLIP (ViT-B/32) and prompts \texttt{"a photo of a \{class\}."}
Support and query images are encoded and L2-normalized; support features serve as cache keys, and one-hot labels as cache values.
Query logits combine zero-shot CLIP logits with a weighted sum over support votes, where the weight depends on the cosine similarity to each support sample. Hyperparameters: $\beta = 5.5$, $\alpha = 1.0$ (default).

\paragraph{InternVL3 Implementation.}
InternVL3-8B~\cite{zhu2025internvl3} combines an InternViT vision encoder with a Qwen2.5-7B language backbone via an MLP connector.
Images are dynamically tiled into up to 4 patches of $448{\times}448$ pixels (aspect-ratio-preserving) and normalized using ImageNet statistics. The model runs in \texttt{bfloat16}; inference uses \texttt{max\_new\_tokens=5} and greedy decoding.

\paragraph{Qwen2.5-VL Implementation.}
We use Qwen2.5-VL-7B-Instruct~\cite{bai2025qwen2}, which combines a dynamic-resolution ViT with 14$\times$14 patch size, window attention, MRoPE, and a Qwen2.5-LM backbone.
Images are processed at native resolution with the processor defaults.
The model runs in \texttt{bfloat16}; inference uses \texttt{max\_new\_tokens=5} and greedy decoding.

\paragraph{Qwen3-VL Implementation.}
We use Qwen3-VL-8B-Instruct~\cite{bai2025qwen3}, which combines a ViT encoder with 16$\times$16 patch size, a Qwen3-LM backbone, Interleaved-MRoPE positional encoding, and DeepStack multi-level feature fusion.
Images are processed with dimensions resized to the nearest multiple of 32.
The model runs in \texttt{bfloat16}; inference uses \texttt{max\_new\_tokens=5} and greedy decoding.



\subsection{Prompt Templates and Formatting}
\paragraph{Logit Scoring Using Confidence.} In Section \ref{sec:eval_ablation}, we discuss the use of the model’s generated confidence score output for \texttt{Yes}. The exact prompt used in the experiment is demonstrated as follows:
\begin{promptbox}
\small
\color{woodbrown}
<Image: $x_{n,k}^s$> \\
<Image: $x^q$> \\
Are the two images depicting the same $D_{\text{info}}$? Reply in exactly this format:\\
ConfidenceYes: <number from 0 to 100>\\
Rules:
ConfidenceYes must be between 0 and 100, where 100 means fully confident they are the same.
Output exactly this line.
\end{promptbox}

\input{ECCV_2026/tab/domain_info_mapping}

\paragraph{Domain Information for Each Dataset.}
Table~\ref{tab:domain_info_mapping} lists the corresponding $D_{\text{info}}$ (domain-specific concept) for all evaluated datasets. For general-purpose datasets such as mini-ImageNet~\cite{vinyals2016matching} and DomainNet~\cite{Peng_2019_ICCV}, we do not specify a dataset-specific domain term. Instead, following the question \texttt{"...depicting the same class?"}, we insert: \texttt{"You should focus on the concept depicted in the} 

\noindent\texttt{image rather than the domain or texture."} on the $D_{\text{info}}$ mode.

\subsection{Instruction Following}
\label{subsec:instruction_following}
\input{ECCV_2026/tab/instruction_following_score}
We observe that the native responses of MLLMs tend to be verbose, which could incur additional noise in answer extraction. Thus, for in-context prompts, we insert instructions to enforce the model to output expected answers. For prompts with semantic labels, we require the model to output options by appending:  \texttt{"Respond with only the option number (1-5), e.g. 1 or 2."} For prompts without semantic labels, we append: \texttt{"Respond with only the class label (e.g. Class 1 or Class 2)."} 

We use a deterministic regex-based parser to extract the predicted class. For semantic-label prompts, we match the first standalone option number; for anonymous-label prompts, we match the first case-insensitive pattern of the form \texttt{Class <n>}. Outputs with no valid match are counted as incorrect.

Furthermore, we compute an instruction-following score, which is the percentage of times the model outputs text that follows the instruction. In Table \ref{tab:instruction_following_score}, Qwen3-VL achieves almost perfect instruction-following performance across all datasets and both prompt settings. InternVL3 also performs well overall, especially in the anonymous setting, but shows lower instruction-following scores than Qwen3-VL on several semantic-label settings.

\newpage

\input{ECCV_2026/tab/cross_model_size}
\subsection{Cross-model Transferability}
To examine whether the gain from decomposition transfers across model scales, we additionally evaluate Qwen3-VL-\textbf{2B/32B} in Table~\ref{tab:model_size}. Decomposition improves the average performance for both model sizes, suggesting that the benefit is not restricted to a particular model capacity. The gains are especially clear on novel datasets and in the anonymous setting, where direct label-based recognition is more challenging. For Qwen3-VL-2B, decomposition substantially improves the anonymous average from 28.7 to 85.4, while for Qwen3-VL-32B, it improves the semantic average from 86.8 to 92.5 and the anonymous average from 80.4 to 87.8. These results indicate that decomposition is robust for improving few-shot recognition across both compact and large MLLMs.

\subsection{Few-shot Action Recognition with Decomposed Prompting}

In this section, we explore the effectiveness of our DeCoDe method adapted to solving the task of few-shot action recognition. We additionally evaluate on UCF101~\cite{soomro2012ucf101} and Diving48~\cite{Li_2018_ECCV}, following the same few-shot evaluation protocol as in the main paper. Diving48 is a fine-grained video action recognition dataset focused on diving actions, making it complementary to UCF101 for evaluating video understanding. We evaluate at 2 FPS for UCF and 8 FPS for Diving48.

\input{ECCV_2026/tab/fsar_table}

As shown in Table~\ref{tab:video_fewshot_results}, Qwen2.5-VL already achieves near-saturated performance on UCF101 in the 0-shot setting, leaving limited room for improvement. In contrast, its performance on Diving48 is much lower, reflecting the fine-grained and temporally sensitive nature of the dataset. Our decision-based variant substantially improves the Diving48 result from 24.7 to 35.5 in the 1-shot setting, while maintaining competitive performance on UCF101. This suggests that our method can be beneficial for few-shot action recognition, while there is a lack of multi-video input training and highly dynamic video training.

\subsection{Exploring Prompts and Domain Information}
In this section, we explore more prompt choices and domain information.

\input{ECCV_2026/tab/dec_label_prompt_variation}

\paragraph{Alternative prompt for the decomposed method with semantic label.} In the original prompt for the decomposed method with semantic label (Figure \ref{fig:prompts}e), we explicitly stated that the first image is from the support set and asked to compare with the support image. To investigate whether this support reference has an effect, we provide an \textbf{alternative prompt}:
\begin{promptbox}
\color{woodbrown}
<Image: $x_{n,k}^s$> \\
<Image: $x^q$> \\
The semantic label of the first image is: $c_n$.\\
Does the second image depict the same $D_{\text{info}}$ as the first image?\\
Answer Yes or No.
\end{promptbox}

In Table \ref{tab:dec_label_prompt_variants}, the performance difference between using the original prompt and the alternative prompt is minor. This result shows that the decomposed prompting method is not sensitive to a slight change in the prompt text. 

\paragraph{Alternative prompt for in-context prompting.} The current 1-shot in-context prompting we use in this work is in an interleaved fashion to start with support images. We explored four alternative prompts: 
\begin{enumerate}
    \item Put the query image first, followed by the support images (\textbf{query first}).
    \item Redefine the few-shot classification problem as an in-context visual matching task (\textbf{visual match}).
    \item Present the support images and query image first, followed by the text description and instruction (\textbf{images then text}).
    \item Standard in-context prompt in Chain of Thought (CoT) style, we use Qwen3-VL-Thinking-8B for this prompt, and set max\_token=600.
\end{enumerate}

\input{ECCV_2026/tab/in_contect_prompt_exploration}

\newpage

\textbf{1. Query first prompt:}
\begin{promptbox}
\color{woodbrown}
<Image: $x^q$> \\
What is this? Match it to one of the options below. \\
<Image: $x_{1,1}^s$> Option 1: $c_1$. \\
...\\
<Image: $x_{5,1}^s$> Option 5: $c_5$. \\
Which option matches the query image shown first? Choose one of: \\
1. $c_1$; ...; 5. $c_5$ 
\end{promptbox}

\textbf{2. Visual match prompt:}
\begin{promptbox}
\color{woodbrown}
<Image: $x_{1,1}^s$> \\
Image 1. \\
...\\
<Image: $x_{5,1}^s$> \\
Image 5. \\
<Image: $x^q$> \\
Which image (1-5) is most visually similar to the last image?\\
Answer with 1-5 only.
\end{promptbox}

\newpage

\textbf{3. Images then text prompt:}
\begin{promptbox}
\color{woodbrown}
<Image: $x_{1,1}^s$> \\
...\\
<Image: $x_{5,1}^s$> \\
<Image: $x^q$> \\
Image 1 belongs to Option 1: $c_1$; ...; Image 5 belongs to Option 5: $c_5$. \\
What class is in the last image? Choose one of the options (1-5). 
\end{promptbox}

\textbf{4. CoT style prompt (Thinking):}
\begin{promptbox}
\color{woodbrown}
<Image: $x_{1,1}^s$> What is this? $c_1$ (option 1). \\
...\\
<Image: $x_{5,1}^s$> What is this? $c_5$ (option 5). \\
The following image is the query image.\\
<Image: $x^q$> \\
So what is this? Choose one of the options: 1. $c_1$; ...; 5. $c_5$ \\
Think step by step, then output exactly one final line in this format: Final answer: <number>
\end{promptbox}

In Table \ref{tab:in_context_prompt_exploration}, under the semantic-label setting, the standard in-context prompt achieves the best average accuracy, while placing the query image first slightly improves on Yoga. Under the anonymous setting, the query-first prompt substantially outperforms the standard in-context prompt, whereas visual-match and images-then-text formulations perform poorly. 

We observe that placing the query image before the support images substantially improves anonymous in-context prompting. One plausible explanation is the autoregressive structure of MLLMs: changing the order may alter how later textual predictions condition on the query and support content. We do not directly test the underlying mechanism, so we treat this as an empirical prompt-order effect rather than a confirmed causal-attention explanation. However, query-first still performs worse than decomposed prompting, which achieves 93.0 average across those three datasets.

\input{ECCV_2026/tab/domain_info_ablation}

\paragraph{Alternative Domain Information.} We experiment with more choices of domain information. We also let the model automatically generate domain information by inputting five randomly sampled images and ask the model to write a short phrase to describe these images. In Table \ref{tab:domain_info_ablation}, the results show that decomposed prompting is fairly robust to the exact choice of $D_{\text{info}}$ as long as the term remains semantically aligned with the dataset. For Yoga, several related variants (e.g., \emph{pose}, \emph{body pose}, and \emph{action}) perform similarly well, with \emph{body pose} giving the best result. For Arabic sign language, the automatically generated term \emph{hand gesture} performs best, even surpassing our manually chosen \emph{sign language}. For Egyptian hieroglyphs, several semantically related terms remain competitive, and the generic term \emph{class} yields the highest accuracy. In contrast, an unrelated descriptor (\emph{kind of flower}) consistently causes a large drop across all datasets. Thus, the precise wording of $D_{\text{info}}$ is not critical, while the semantic domain relevance is essential.

\input{ECCV_2026/tab/baseline_with_dinfo_table}
\paragraph{Applying domain information to in-context prompting.}
We also test whether $D_{\text{info}}$ alone improves standard in-context prompting. Specifically, we modify the query instruction from ``So what is this?'' to ``So what is this $D_{\text{info}}$?'' while keeping the same N-way in-context structure. As shown in Table~\ref{tab:baseline_Dinfo}, this change yields only marginal average gains for direct in-context inference, from 80.3 to 80.8 in the semantic setting and from 27.8 to 28.4 in the anonymous setting. In contrast, $D_{\text{info}}$ provides much larger gains when combined with decomposed prompting. This suggests that domain information is most useful when it guides pairwise support--query comparison, rather than when it is simply inserted into the final N-way question.

\begin{figure}[!t]
\centering
\includegraphics[width=0.99\linewidth]{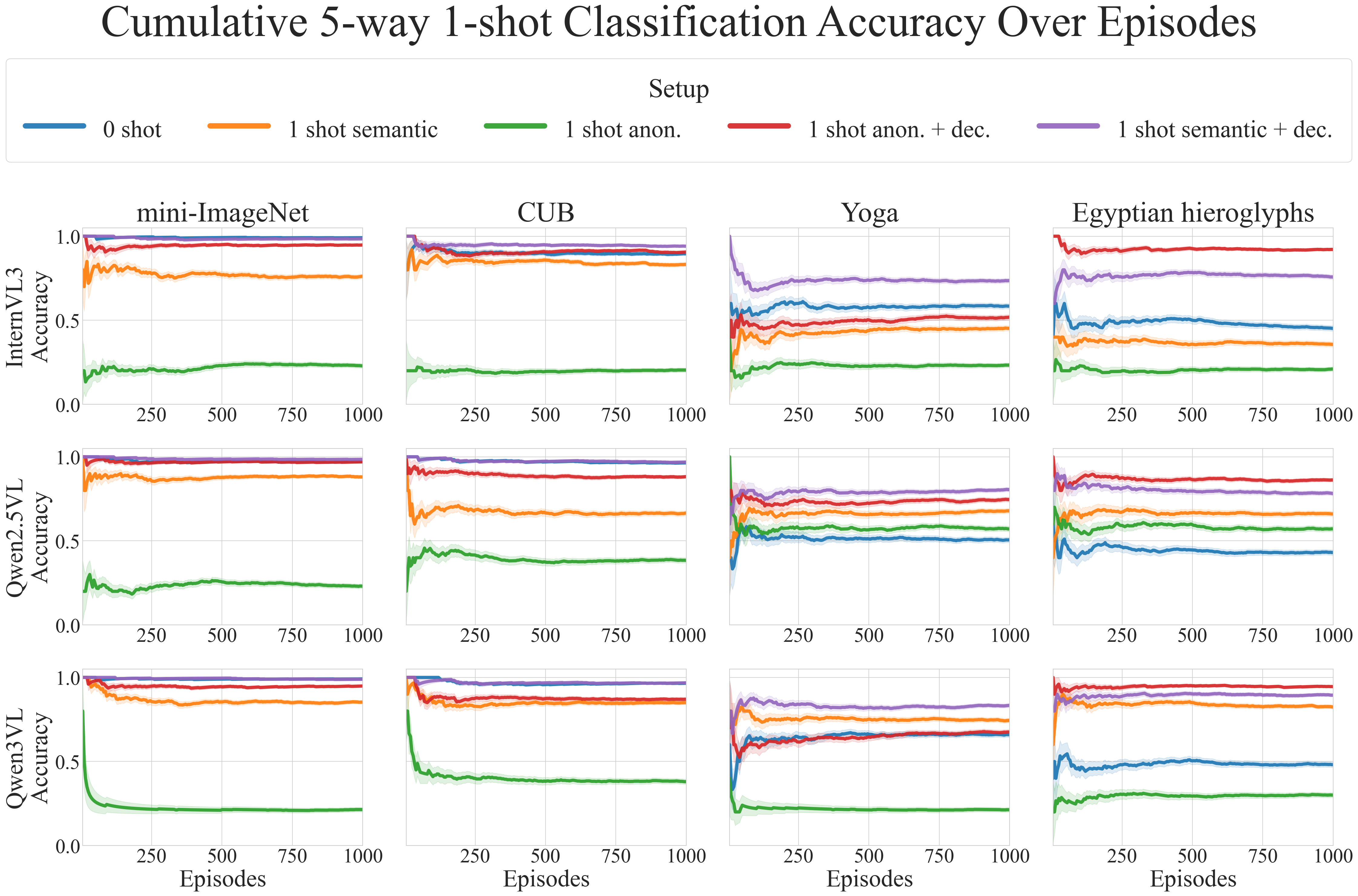}
\caption{
Cumulative 5-way 1-shot episodic classification accuracy across four datasets for three MLLMs. The x-axis shows evaluated episodes (up to 1000; 5 episodes per logging step), and the y-axis shows cumulative accuracy. Solid lines correspond to prompting setups, where semantic denotes using semantic labels, anon. denotes removing semantic labels, and dec. denotes decomposed prompting (0 shot, 1 shot semantic, 1 shot anon., 1 shot anon. + dec., 1 shot semantic + dec.). Shaded regions indicate ±1 standard error around each cumulative accuracy curve.
}
\label{fig:episode_curves_se_band}
\end{figure}

\subsection{Episode Curves with Standard Error}
\label{subsec:episode_curve}

We report cumulative episodic accuracy with uncertainty bands to verify how many episodes are sufficient for reliable 5-way 0-shot/1-shot accuracy. This analysis is important because few-shot episodic results can be noisy at small sample sizes; without uncertainty, early fluctuations can be over-interpreted.

To address this, as shown in Figure \ref{fig:episode_curves_se_band}, we run each model on 1000 episodes per dataset (logged every 5 episodes), plot cumulative accuracy trajectories, and overlay standard-error bands at each point. This directly shows both performance trends and estimated stability as more episodes are accumulated. Across all models and datasets, we observe a common pattern: fast early changes, progressively narrower uncertainty, and eventual flattening of the curves. The relative ordering of prompting variants is largely stable after the initial transient stage, and by around 1000 episodes, the trajectories are near-stationary with tight uncertainty bands, indicating effective convergence.

\subsection{Logit Distribution for Scoring Tokens}

\begin{figure}[!htbp]
\centering
\includegraphics[width=0.95\linewidth]{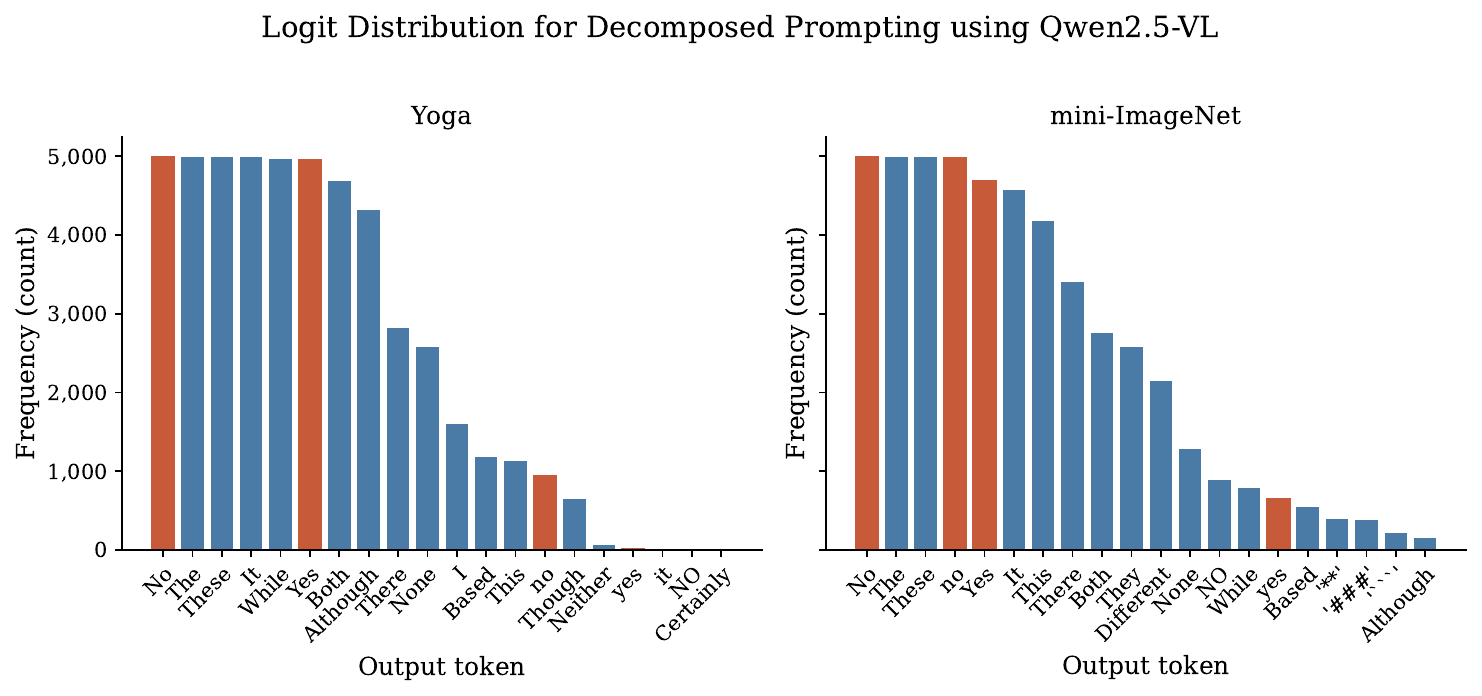}
\caption{
Logit distribution for decomposed prompting using Qwen2.5-VL on Yoga (left) and Mini-ImageNet (right). For each decomposed inference in each 5-way 1-shot episode, we collect the top-10 predicted tokens over all support–query comparisons. Bars show token frequency; Yes/No are highlighted as the intended answer tokens.
}
\label{fig:logit_distribution}
\end{figure}

To investigate the logit choice of \texttt{Yes} for the scoring rule of decomposed prompting, we analyze the output logit distribution if we change the original instruction \texttt{"Answer Yes or No."} to free-form answer format: \texttt{"Answer: "}. We run Qwen2.5-VL on Yoga \cite{kaggle_yoga} and Mini-ImageNet \cite{vinyals2016matching} for 1000 episodes and record the top ten logits by their logit score in each episode. Each episode corresponds to five inferences because of decomposition. Figure \ref{fig:logit_distribution} shows that tokens such as \texttt{Yes} and \texttt{No} frequently appear among high-probability candidates, which is consistent with using the \texttt{Yes} logit as a simple scoring signal. We view this analysis as supportive rather than definitive: it provides intuition for the design choice, but does not by itself prove that \texttt{Yes} is the optimal token for all settings.

\subsection{Failure Cases of DeCoDe}

\begin{figure}[t!]
\centering
\includegraphics[width=0.98\linewidth]{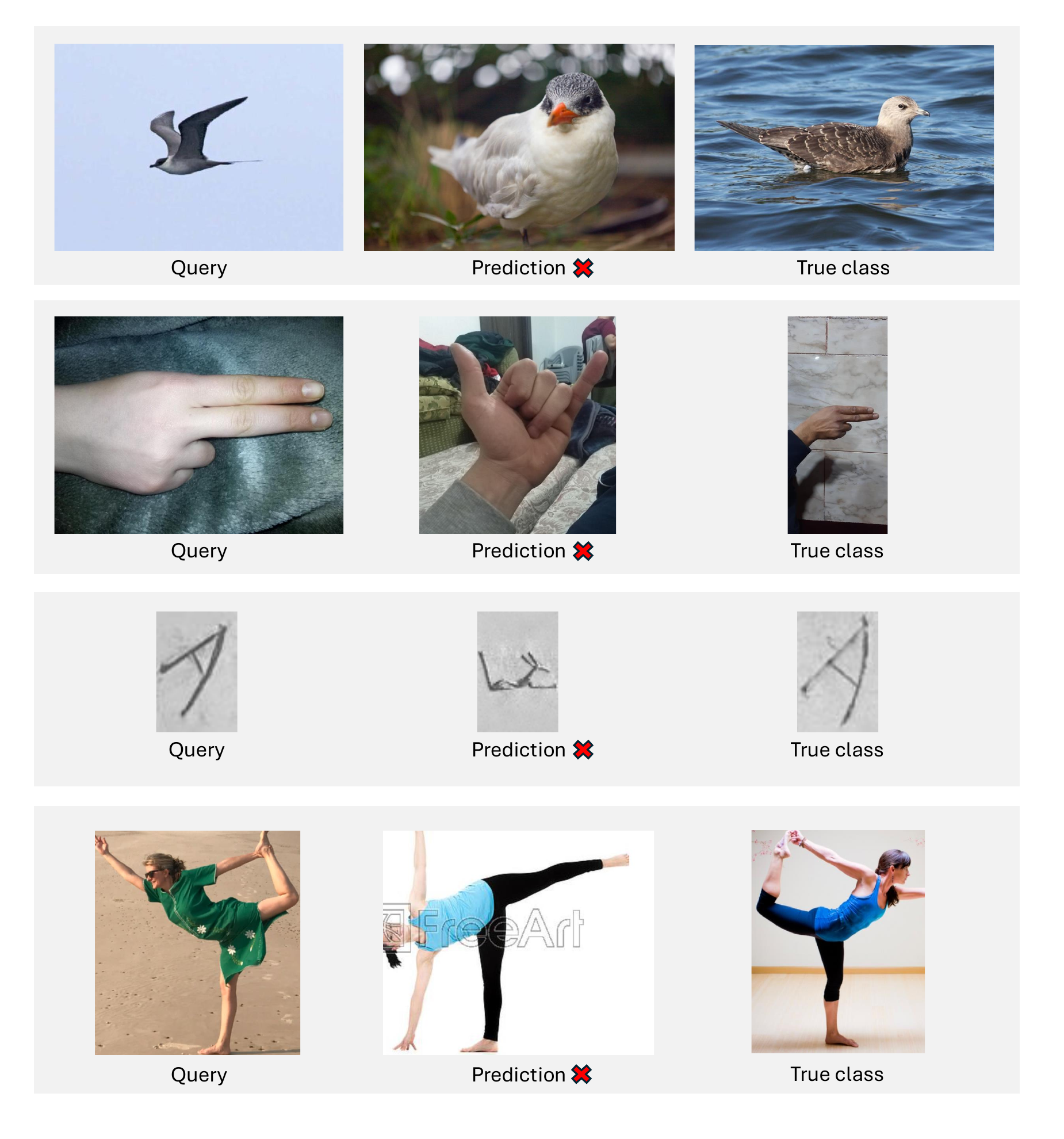}
\caption{
Failure Cases of our DeCoDe method with labels and without $D_\text{info}$. 
}
\label{fig:failure_cases}
\end{figure}

We further provide failure cases of our DeCoDe method with labels and without domain information in Figure \ref {fig:failure_cases}. 

%% file: ECCV_2026/tab/dataset_details.tex
\begin{table}[H]
\centering
\begin{tabular}{l|c|c|c|c}
\toprule
\textbf{Dataset} & \textbf{Classes} & \textbf{Total Images} & \textbf{Avg Img/Cls} & \textbf{Eval Classes} \\
\midrule
\multicolumn{5}{l}{\cellcolor{sotaStdBlue}\scalebox{0.8}Standard} \\
mini-ImageNet~\cite{vinyals2016matching}       & 100 & 60,000  & 600   & 20 \\
UCF101~\cite{soomro2012ucf101}             & 101 & 13,320  & 132   & 21 \\
CUB~\cite{wah2011caltech}                & 200 & 11,788  & 59    & 50 \\
Aircraft~\cite{maji13fine-grained}      & 100 & 10,000  & 100   & 100 \\
Dogs~\cite{khosla2011novel}               & 120 & 20,580  & 172   & 30 \\
DomainNet~\cite{Peng_2019_ICCV}          & 345 & 596,006 & 1,728 & 61 \\
\midrule
\multicolumn{5}{l}{\cellcolor{sotaNovelAmber}\scalebox{0.8}Novel} \\
Lego bricks \cite{kaggle_lego}            & 20  & 4,580   & 229    & 20 \\
Industrial \cite{kaggle_industrial}       & 10  & 100,000 & 10,000 & 10 \\
Yoga pose \cite{kaggle_yoga}                  & 107 & 5,991   & 56     & 107 \\
Egyptian hieroglyphs \cite{franken2013automatic}  & 164 & 17,388  & 106    & 164 \\
Flying insects \cite{kaggle_insect}         & 100 & 13,594  & 136    & 100 \\
Arabic sign language \cite{https://doi.org/10.48550/arxiv.2301.11932}   & 31  & 7,856   & 253    & 31 \\
\bottomrule
\end{tabular}
\vspace{2mm}
\caption{Overview of few-shot learning datasets. ``Eval Classes'' is the number of classes used for evaluation (sampled test split).}
\label{tab:dataset_details}
\end{table}

%% file: ECCV_2026/tab/domain_info_mapping.tex
\begin{table}[t]
\centering
\scriptsize
\setlength{\tabcolsep}{2.8pt}
\renewcommand{\arraystretch}{1.05}

\begin{tabular}{
>{\raggedright\arraybackslash}p{0.24\linewidth}|
>{\raggedright\arraybackslash}p{0.20\linewidth}|
>{\raggedright\arraybackslash}p{0.43\linewidth}
}
\toprule
\textbf{Dataset} & \textbf{$D_{\text{info}}$ text} & \textbf{Example class labels} \\
\midrule

mini-ImageNet~\cite{vinyals2016matching}
& --
& house finch; triceratops; Tibetan mastiff \\

UCF101~\cite{soomro2012ucf101}
& action category
& ApplyEyeMakeup; BasketballDunk; PlayingGuitar \\

CUB~\cite{wah2011caltech}
& bird species
& Black-footed Albatross; Laysan Albatross; Fish Crow \\

Aircraft~\cite{maji13fine-grained}
& aircraft variant
& Gulfstream V; 727-200; Saab 340 \\

Dogs~\cite{khosla2011novel}
& dog breed
& Chihuahua; Saluki; Pekinese \\

DomainNet~\cite{Peng_2019_ICCV}
& --
& aircraft carrier; cello; lighthouse \\

Lego~\cite{kaggle_lego}
& Lego brick type
& Brick 1$\times$3; Brick 2$\times$2 Slope; Plate 2$\times$4 \\

Industrial~\cite{kaggle_industrial}
& industrial product
& bracket big; engine part cooler square; screw \\

Yoga~\cite{kaggle_yoga}
& yoga pose
& tittibhasana; vajrasana; bakasana \\

Egyptian hieroglyph~\cite{franken2013automatic}
& Egyptian hieroglyph
& road with shrubs; cow with calf; sun disk \\

Flying insects~\cite{kaggle_insect}
& insect species
& Red Cracker; An 88; Danaid Eggfly \\

Arabic sign language~\cite{https://doi.org/10.48550/arxiv.2301.11932}
& sign language
& Alef; Reh; Qaf \\

\bottomrule
\end{tabular}

\vspace{2mm}
\caption{Mapping of datasets to their domain information and example class labels.}
\label{tab:domain_info_mapping}
\end{table}

%% file: ECCV_2026/tab/instruction_following_score.tex





{
\hypersetup{hidelinks}

\begin{table}[t]
\centering
\begingroup
\small
\setlength{\tabcolsep}{4.2pt}
\renewcommand{\arraystretch}{1.03}
\begin{tabular}{l cc cc cc cc c}
\toprule
\multicolumn{1}{c}{} & \multicolumn{9}{c}{\textbf{Instruction-following score (\%) $\uparrow$}} \\
\cmidrule(lr){2-10}
\textbf{Model}
& \multicolumn{2}{c}{\textbf{mini}} 
& \multicolumn{2}{c}{\textbf{CUB}}
& \multicolumn{2}{c}{\textbf{Yoga}}
& \multicolumn{2}{c}{\textbf{Lego}}
& \textbf{Avg.} \\
\cmidrule(lr){2-3}\cmidrule(lr){4-5}\cmidrule(lr){6-7}\cmidrule(lr){8-9}
& \textbf{Sem.} & \textbf{Anon.}
& \textbf{Sem.} & \textbf{Anon.}
& \textbf{Sem.} & \textbf{Anon.}
& \textbf{Sem.} & \textbf{Anon.}
&  \\
\midrule
Qwen3-VL  & 100.0 & 100.0 & 100.0 & 100.0 & 99.9 & 100.0 & 99.1 & 100.0 & 99.9 \\
InternVL3 & 87.3  & 100.0 & 94.0  & 100.0 & 89.6 & 99.2  & 100.0 & 100.0 & 96.3 \\
\bottomrule
\end{tabular}
\endgroup
\vspace{2mm}
\caption{Instruction-following score (\%) of in-context prompting on four datasets for both semantic-label (\textbf{Sem.}) and anonymous (\textbf{Anon.}) settings, computed as $100 -$ instruction-following error rate.}
\label{tab:instruction_following_score}
\end{table}
}

%% file: ECCV_2026/tab/cross_model_size.tex
\begin{table}[ht]
    \footnotesize
    \centering
    \setlength{\tabcolsep}{3pt}
    \resizebox{0.8\linewidth}{!}{%
    \begin{tabular}{ll|ccc|ccc}
    \toprule
    & & \multicolumn{3}{c}{Qwen3-VL 2B} & \multicolumn{3}{c}{Qwen3-VL 32B} \\
    \toprule
    \multicolumn{2}{l}{Method} & std (m,c,d) & nvl (y,h,s) & Avg  & std (m,c,d) & nvl (y,h,s) & Avg  \\
    \midrule
    & 0-shot & \textbf{92.7} & 35.2 & 63.9   & 97.3 & 53.7 & 75.5  \\
    & 1-shot & 85.6 & 60.2 & 72.8   & 91.7 & 81.9 & 86.8  \\
    & \cellcolor{almond}1-shot dec. & \cellcolor{almond}84.8 & \cellcolor{almond}\textbf{68.6} & \cellcolor{almond}\textbf{76.7}  & \cellcolor{almond}\textbf{97.6} & \cellcolor{almond}\textbf{87.3} & \cellcolor{almond}\textbf{92.5}  \\
    \midrule
    \multicolumn{8}{l}{\cellcolor{sotaModelGrey}{Above: with labels (Semantic); Below: without labels (Anonymous)}} \\[-1pt]
    \midrule
    & 1-shot & 25.9 & 31.6 & 28.7  & 75.2 & 85.6 & 80.4   \\
    & \cellcolor{almond}1-shot dec. & \cellcolor{almond}\textbf{93.3} & \cellcolor{almond}\textbf{77.5} & \cellcolor{almond}\textbf{85.4}  & \cellcolor{almond}\textbf{86.1} & \cellcolor{almond}\textbf{89.5} & \cellcolor{almond}\textbf{87.8}   \\
    \bottomrule
    \end{tabular}
    }
    \vspace{2mm}
    \caption{Evaluation on Qwen3-VL 2B and 32B models}
    \label{tab:model_size}
\end{table}

%% file: ECCV_2026/tab/fsar_table.tex
\begin{wraptable}[10]{r}{0.46\linewidth}
\vspace{-5mm}
\centering
\scriptsize
\setlength{\tabcolsep}{3.2pt}
\renewcommand{\arraystretch}{0.94}

\caption{Video few-shot results on UCF and Diving48.}
\label{tab:video_fewshot_results}

\begin{tabular}{lcc}
\toprule
\textbf{Setting} & \textbf{UCF101} & \textbf{Diving48} \\
\midrule
\multicolumn{3}{l}{\cellcolor{sotaModelGrey}\textbf{With labels (Semantic)}} \\[-1pt]
\midrule
\multicolumn{3}{l}{\cellcolor{sotaModelTint}\textit{Qwen2.5-VL-7B}} \\
0-shot      & 95.9 & 22.7 \\
1-shot      & 96.8 & 24.7 \\
\rowcolor{almond}
1-shot dec. & 96.4 & 35.5 \\
\bottomrule
\end{tabular}

\end{wraptable}

%% file: ECCV_2026/tab/dec_label_prompt_variation.tex
\begin{table}[t]
\centering
\small
\setlength{\tabcolsep}{5pt}
\renewcommand{\arraystretch}{1.05}
\begin{tabular}{l|c|c|c|c|c|c|c}
\toprule
\textbf{Prompt} & \textbf{mini} & \textbf{CUB} & \textbf{Dogs} & \textbf{Lego} & \textbf{Yoga} & \textbf{Hiero.} & \textbf{Avg} \\
\midrule
Original prompt    & 99.0 & 97.9 & 96.6 & 72.9 & 88.3 & 90.8 & 90.9 \\
\makecell[l]{Alternative prompt \\ (without support reference)} & 98.7 & 97.3 & 96.3 & 73.1 & 89.5 & 88.8 & 90.6 \\
\bottomrule
\end{tabular}
\vspace{1mm}
\caption{Few-shot classification accuracy (\%) of Qwen3-VL using decomposition with semantic label and domain information under two prompt variants.}
\label{tab:dec_label_prompt_variants}
\end{table}

%% file: ECCV_2026/tab/in_contect_prompt_exploration.tex
{
\hypersetup{hidelinks}

\providecolor{sotaNovelAmber}{RGB}{255,210,160}
\providecolor{sotaModelGrey}{RGB}{232,232,232}

\begin{table}[t!]
\centering
\small
\begingroup
\setlength{\tabcolsep}{5pt}
\renewcommand{\arraystretch}{1.02}
\begin{tabular}{lccc|c}
\toprule
\multicolumn{1}{l}{} &
\multicolumn{4}{c}{\cellcolor{sotaNovelAmber}\scalebox{0.7}{\textbf{Novel Datasets}}} \\
\cmidrule(lr){2-5}
\textbf{Prompt Setting}
& \textbf{Yoga} & \textbf{Hiero.} & \textbf{Sign} & \textbf{Avg.} \\

\midrule
\multicolumn{5}{l}{\cellcolor{sotaModelGrey}\textbf{With semantic label}} \\[-1pt]
\midrule

Standard in-context & 74.5 & \textbf{82.4} & \textbf{68.4} & \textbf{75.1} \\
1. Query first      & \textbf{76.7} & 77.8 & 56.1 & 70.2 \\
3. Images then text & 9.0 & 13.8 & 12.5 & 11.8 \\
4. CoT (Thinking)   & 41.7 & 68.4 & 24.1 & 44.7 \\[2pt]

\midrule
\multicolumn{5}{l}{\cellcolor{sotaModelGrey}\textbf{Anonymous}} \\[-1pt]
\midrule

Standard in-context & 20.3 & 30.0 & 28.1 & 26.1 \\
1. Query first      & \textbf{70.5} & \textbf{80.5} & \textbf{52.5} & \textbf{67.8} \\
2. Visual match     & 18.5 & 12.5 & 15.2 & 15.4 \\
3. Images then text & 5.9 & 20.8 & 8.4 & 11.7 \\
4. CoT (Thinking)   & 9.0 & 47.3 & 2.3 & 19.5 \\

\bottomrule
\end{tabular}
\endgroup
\vspace{2mm}
\caption{In-context prompt exploration on three novel datasets using Qwen3-VL. Standard in-context denotes the interleaved in-context prompting used in the main paper. We experimented with both the semantic and anonymous settings.}
\label{tab:in_context_prompt_exploration}
\end{table}
}

%% file: ECCV_2026/tab/domain_info_ablation.tex
{
\hypersetup{hidelinks}

\begin{table}[t!]
\centering
\begingroup
\scriptsize
\setlength{\tabcolsep}{2.8pt}
\renewcommand{\arraystretch}{1.0}
\begin{adjustbox}{max width=\linewidth}
\begin{tabular}{l c l c l c}
\toprule
\multicolumn{2}{c}{\textbf{Yoga}} &
\multicolumn{2}{c}{\textbf{Sign}} &
\multicolumn{2}{c}{\textbf{Hiero.}} \\
\cmidrule(lr){1-2}\cmidrule(lr){3-4}\cmidrule(lr){5-6}
\textbf{$D_{\text{info}}$} & \textbf{Acc.} &
\textbf{$D_{\text{info}}$} & \textbf{Acc.} &
\textbf{$D_{\text{info}}$} & \textbf{Acc.} \\
\midrule
class                       & 67.5 & class                         & 72.6 & class                        & \textbf{94.5} \\
\rowcolor{almond} yoga pose \textcolor{gray}{(we use)}          & 89.6 & sign language \textcolor{gray}{(we use)}        & 86.6 & Egyptian hieroglyph \textcolor{gray}{(we use)} & 91.2 \\
pose                        & 89.9 & Arabic sign language          & 66.4 & hieroglyph                   & 93.7 \\
body pose                   & \textbf{90.8} & Arabic alphabet sign language & 59.6 & drawing                      & 94.0 \\
action                      & \underline{90.5} & Arabic sign language alphabet & 57.1 & icon                         & \underline{94.2} \\
\rowcolor{lightgreenrow} yoga stretch \textcolor{gray}{(auto)}         & 88.8 & hand gesture \textcolor{gray}{(auto)}           & \textbf{90.0} & ancient symbol \textcolor{gray}{(auto)}        & 93.5 \\
\rowcolor{lightredrow} kind of flower \textcolor{gray}{(unrelated)}  & 20.6 & kind of flower \textcolor{gray}{(unrelated)}    & 24.0 & kind of flower \textcolor{gray}{(unrelated)}   & 72.9 \\
\bottomrule
\end{tabular}
\end{adjustbox}
\endgroup
\vspace{2mm}
\caption{Effect of different domain information $D_{\text{info}}$ in decomposed prompting (anonymous) with Qwen3-VL on three novel datasets. ``(we use)'' denotes the domain term used in the main paper, while ``(auto)'' denotes a term automatically generated by Qwen3-VL. Best and second-best results are highlighted in \textbf{bold} and \underline{underlined}.}
\label{tab:domain_info_ablation}
\end{table}
}

%% file: ECCV_2026/tab/baseline_with_dinfo_table.tex
\begin{table}[t!]
    \footnotesize
    \centering
    \setlength{\tabcolsep}{3pt}
    \resizebox{0.9\linewidth}{!}{%
    \begin{tabular}{lccc}
    \toprule
    Qwen3-VL-8B & std (mini, cub, dog) & nvl (yoga, hiero., sign) & Avg  \\
    \midrule
    1-shot \textcolor{gray}{(baseline)} & 85.5 & 75.1 & 80.3  \\
    1-shot +$D_{\text{info}}$ & 87.5 & 74.1 & 80.8  \\
    \rowcolor{almond}1-shot dec. \textcolor{gray}{(ours)} & 97.3 & 81.2 & 89.2  \\
    \rowcolor{almond}1-shot dec. +$D_{\text{info}}$ \textcolor{gray}{(ours)} & \textbf{97.8} & \textbf{86.8} & \textbf{92.3} \\
    \multicolumn{4}{l}{\cellcolor{sotaModelGrey}Above: with labels (Semantic); Below: without labels (Anonymous)} \\[-1pt]
    \midrule
    1-shot \textcolor{gray}{(baseline)} & 29.0 & 26.5 & 27.8  \\
    1-shot +$D_{\text{info}}$ & 31.9 & 24.8 & 28.4  \\
    \rowcolor{almond}1-shot dec. \textcolor{gray}{(ours)} & 90.0 & 78.2 & 84.1  \\
    \rowcolor{almond}1-shot dec. +$D_{\text{info}}$ \textcolor{gray}{(ours)} & \textbf{95.8} & \textbf{89.1} & \textbf{92.5} \\
    \bottomrule
    \end{tabular}
    }
    \vspace{2mm}
    \caption{Effect of applying $D_{\text{info}}$ to in-context and decomposed prompting.}
    \label{tab:baseline_Dinfo}
\end{table}

%% file: ECCV_2026/bib/longstrings.bib
@String(IJCV  = {Int. J. Comput. Vis.})

@String(CVPR  = {IEEE Conf. Comput. Vis. Pattern Recog.})

@String(ICCV  = {Int. Conf. Comput. Vis.})

@String(ECCV  = {Eur. Conf. Comput. Vis.})

@String(ICML  = {Int. Conf. Mach. Learn.})

@String(ICLR  = {Int. Conf. Learn. Represent.})

@String(AAAI  = {AAAI})

@String(ACMMM = {ACM Int. Conf. Multimedia})

@String(IJCV  = {IJCV})

@String(CVPR  = {CVPR})

@String(ICCV  = {ICCV})

@String(ECCV  = {ECCV})

@String(ICML  = {ICML})

@String(ICLR  = {ICLR})

@String(ACMMM = {ACM MM})


%% file: ECCV_2026/bib/main.bib
@String(IJCV = {Int. J. Comput. Vis.})

@String(CVPR= {IEEE Conf. Comput. Vis. Pattern Recog.})

@String(ICCV= {Int. Conf. Comput. Vis.})

@String(ECCV= {Eur. Conf. Comput. Vis.})

@String(NIPS= {Adv. Neural Inform. Process. Syst.})

@String(ACMMM= {ACM Int. Conf. Multimedia})

@String(ICLR = {Int. Conf. Learn. Represent.})

@String(AAAI = {AAAI})

@String(NIPS  = {NeurIPS})

@inproceedings{recyclelora,
      title={Unlocking Tuning-Free Few-Shot Adaptability in Visual Foundation Models by Recycling Pre-Tuned LoRAs}, 
      author={Zixuan Hu and Yongxian Wei and Li Shen and Chun Yuan and Dacheng Tao},
      year={2025},
      booktitle=CVPR
}

@inproceedings{sav,
      title={Enhancing Few-Shot Vision-Language Classification with Large Multimodal Model Features}, 
      author={Chancharik Mitra and Brandon Huang and Tianning Chai and Zhiqiu Lin and Assaf Arbelle and Rogerio Feris and Leonid Karlinsky and Trevor Darrell and Deva Ramanan and Roei Herzig},
      year={2025},
      booktitle=ICCV
}

@inproceedings{proker,
      title={ProKeR: A Kernel Perspective on Few-Shot Adaptation of Large Vision-Language Models}, 
      author={Yassir Bendou and Amine Ouasfi and Vincent Gripon and Adnane Boukhayma},
      year={2025},
      booktitle=CVPR

}

@inproceedings{ldc,
      title={Logits DeConfusion with CLIP for Few-Shot Learning}, 
      author={Shuo Li and Fang Liu and Zehua Hao and Xinyi Wang and Lingling Li and Xu Liu and Puhua Chen and Wenping Ma},
      year={2025},
      booktitle=CVPR
}

@inproceedings{2sfs,
      title={Rethinking Few-Shot Adaptation of Vision-Language Models in Two Stages}, 
      author={Matteo Farina and Massimiliano Mancini and Giovanni Iacca and Elisa Ricci},
      year={2025},
      booktitle=CVPR
}

@article{coop,
   title={Learning to Prompt for Vision-Language Models},
   author={Zhou, Kaiyang and Yang, Jingkang and Loy, Chen Change and Liu, Ziwei},
   journal=IJCV,
   year={2022},
}

@inproceedings{coopop,
      title={Conditional Prompt Learning for Vision-Language Models}, 
      author={Kaiyang Zhou and Jingkang Yang and Chen Change Loy and Ziwei Liu},
      year={2022},
      booktitle=CVPR
}

@inproceedings{prograd,
      title={Prompt-aligned Gradient for Prompt Tuning}, 
      author={Beier Zhu and Yulei Niu and Yucheng Han and Yue Wu and Hanwang Zhang},
      year={2023},
      booktitle=ICCV
}

@inproceedings{kgcoop,
      title={Visual-Language Prompt Tuning with Knowledge-guided Context Optimization}, 
      author={Hantao Yao and Rui Zhang and Changsheng Xu},
      year={2023},
      booktitle=CVPR
}

@inproceedings{maple,
      title={MaPLe: Multi-modal Prompt Learning}, 
      author={Muhammad Uzair Khattak and Hanoona Rasheed and Muhammad Maaz and Salman Khan and Fahad Shahbaz Khan},
      year={2023},
      booktitle=CVPR
}

@inproceedings{plot,
      title={PLOT: Prompt Learning with Optimal Transport for Vision-Language Models}, 
      author={Guangyi Chen and Weiran Yao and Xiangchen Song and Xinyue Li and Yongming Rao and Kun Zhang},
      year={2023},
      booktitle=ICLR
}

@article{clipadapter,
      title={CLIP-Adapter: Better Vision-Language Models with Feature Adapters}, 
      author={Peng Gao and Shijie Geng and Renrui Zhang and Teli Ma and Rongyao Fang and Yongfeng Zhang and Hongsheng Li and Yu Qiao},
      year={2024},
      journal=IJCV
}

@inproceedings{taskres,
      title={Task Residual for Tuning Vision-Language Models}, 
      author={Tao Yu and Zhihe Lu and Xin Jin and Zhibo Chen and Xinchao Wang},
      year={2023},
      booktitle=CVPR
}

@inproceedings{tipadapter,
      title={Tip-Adapter: Training-free Adaption of CLIP for Few-shot Classification}, 
      author={Renrui Zhang and Zhang Wei and Rongyao Fang and Peng Gao and Kunchang Li and Jifeng Dai and Yu Qiao and Hongsheng Li},
      year={2022},
      booktitle=ECCV
}

@inproceedings{susx,
      title={SuS-X: Training-Free Name-Only Transfer of Vision-Language Models}, 
      author={Vishaal Udandarao and Ankush Gupta and Samuel Albanie},
      year={2023},
      booktitle=ICCV
}

@inproceedings{lpclip,
      title={Learning Transferable Visual Models From Natural Language Supervision}, 
      author={Alec Radford and Jong Wook Kim and Chris Hallacy and Aditya Ramesh and Gabriel Goh and Sandhini Agarwal and Girish Sastry and Amanda Askell and Pamela Mishkin and Jack Clark and Gretchen Krueger and Ilya Sutskever},
      year={2021},
      booktitle=ICML
}

@inproceedings{zhai2023sigmoid,
      title={Sigmoid loss for language image pre-training},
      author={Zhai, Xiaohua and Mustafa, Basil and Kolesnikov, Alexander and Beyer, Lucas},
      booktitle=ICCV,
      year={2023}
}

@article{oquab2023dinov2,
      title={Dinov2: Learning robust visual features without supervision},
      author={Oquab, Maxime and Darcet, Timoth{\'e}e and Moutakanni, Th{\'e}o and Vo, Huy and Szafraniec, Marc and Khalidov, Vasil and Fernandez, Pierre and Haziza, Daniel and Massa, Francisco and El-Nouby, Alaaeldin and others},
      journal={arXiv preprint arXiv:2304.07193},
      year={2023}
}

@inproceedings{liu2024,
      author = {Liu, Fan and Cai, Wenwen and Huo, Jian and Zhang, Chuanyi and Chen, Delong and Zhou, Jun},
      title = {Making large vision language models to be good few-shot learners},
      year = {2025},
      booktitle=AAAI
}

@inproceedings{baldassini2024,
      title={What Makes Multimodal In-Context Learning Work?}, 
      author={Folco Bertini Baldassini and Mustafa Shukor and Matthieu Cord and Laure Soulier and Benjamin Piwowarski},
      year={2024},
      booktitle=CVPR
}

@inproceedings{zhao2021calibrate,
      title={Calibrate before use: Improving few-shot performance of language models},
      author={Zhao, Zihao and Wallace, Eric and Feng, Shi and Klein, Dan and Singh, Sameer},
      booktitle=ICML,
      year={2021}
}

@article{bai2025qwen3,
      title={Qwen3-vl technical report},
      author={Bai, Shuai and Cai, Yuxuan and Chen, Ruizhe and Chen, Keqin and Chen, Xionghui and Cheng, Zesen and Deng, Lianghao and Ding, Wei and Gao, Chang and Ge, Chunjiang and others},
      journal={arXiv preprint arXiv:2511.21631},
      year={2025}
}

@article{bai2025qwen2,
      title={Qwen2.5-VL Technical Report},
      author={Bai, Shuai and Chen, Keqin and Liu, Xuejing and Wang, Jialin and Ge, Wenbin and Song, Sibo and Dang, Kai and Wang, Peng and Wang, Shijie and Tang, Jun and others},
      journal={arXiv preprint arXiv:2502.13923},
      year={2025}
}

@article{zhu2025internvl3,
      title={Internvl3: Exploring advanced training and test-time recipes for open-source multimodal models},
      author={Zhu, Jinguo and Wang, Weiyun and Chen, Zhe and Liu, Zhaoyang and Ye, Shenglong and Gu, Lixin and Tian, Hao and Duan, Yuchen and Su, Weijie and Shao, Jie and others},
      journal={arXiv preprint arXiv:2504.10479},
      year={2025}
}

@inproceedings{vinyals2016matching,
      title={Matching networks for one shot learning},
      author={Vinyals, Oriol and Blundell, Charles and Lillicrap, Timothy and Wierstra, Daan and others},
      booktitle=NIPS,
      year={2016}
}

@techreport{wah2011caltech,
      title={The caltech-ucsd birds-200-2011 dataset},
      author={Wah, Catherine and Branson, Steve and Welinder, Peter and Perona, Pietro and Belongie, Serge and others}
}

@article{maji13fine-grained,
       title={Fine-Grained Visual Classification of Aircraft},
       author={S. Maji and J. Kannala and E. Rahtu and M. Blaschko and A. Vedaldi},
       year={2013},
       journal={arXiv preprint arXiv:1306.5151}
}

@inproceedings{lin2024evaluating,
  title={Evaluating text-to-visual generation with image-to-text generation},
  author={Lin, Zhiqiu and Pathak, Deepak and Li, Baiqi and Li, Jiayao and Xia, Xide and Neubig, Graham and Zhang, Pengchuan and Ramanan, Deva},
  booktitle=ECCV,
  year={2024},
}

@inproceedings{khosla2011novel,
  title={Novel dataset for fine-grained image categorization: Stanford dogs},
  author={Khosla, Aditya and Jayadevaprakash, Nityananda and Yao, Bangpeng and Li, Fei-Fei}
}

@article{soomro2012ucf101,
  title={Ucf101: A dataset of 101 human actions classes from videos in the wild},
  author={Soomro, Khurram and Zamir, Amir Roshan and Shah, Mubarak},
  journal={arXiv preprint arXiv:1212.0402},
  year={2012}
}

@inproceedings{Peng_2019_ICCV,
author = {Peng, Xingchao and Bai, Qinxun and Xia, Xide and Huang, Zijun and Saenko, Kate and Wang, Bo},
title = {Moment Matching for Multi-Source Domain Adaptation},
booktitle =ICCV,
year = {2019}
}

@inproceedings{shvetsova2025unbiasing,
  title={Unbiasing through textual descriptions: Mitigating representation bias in video benchmarks},
  author={Shvetsova, Nina and Nagrani, Arsha and Schiele, Bernt and Kuehne, Hilde and Rupprecht, Christian},
  booktitle=CVPR,
  year={2025}
}

@inproceedings{hu2022lora,
  title={LoRA: Low-Rank Adaptation of Large Language Models},
  author={Hu, Edward J and Shen, Yelong and Wallis, Phillip and Allen-Zhu, Zeyuan and Li, Yuanzhi and Wang, Shentan and Wang, Lu and Chen, Weizhu},
  booktitle=ICLR,
  year={2022}
}

@misc{kaggle_industrial,
  author       = {Schuerrle, Berit and Sankarappan, Vekatesh},
  title        = {Industrial Classification Dataset},
  year         = {2023},
  note         = {Kaggle}
}

@misc{kaggle_yoga,
  author       = {Saxena, Shruti},
  title        = {Yoga Pose Image Classification Dataset},
  year         = {2021},
  note         = {Kaggle}
}

@misc{kaggle_lego,
  author       = {Garciam, Paco},
  title        = {Lego Brick Sorting Image Recognition},
  year         = {2019},
  note         = {Kaggle}
}

@misc{kaggle_insect,
  author       = {Piosenka, Gerry},
  title        = {Butterfly and Moths Image Classification 100 species},
  year         = {2023},
  note         = {Kaggle}
}

@article{https://doi.org/10.48550/arxiv.2301.11932,
  author = {Al-Barham, Muhammad and Alsharkawi, Adham and Al-Yaman, Musa and Al-Fetyani, Mohammad and Elnagar, Ashraf and SaAleek, Ahmad Abu and Al-Odat, Mohammad},
  title = {RGB Arabic Alphabets Sign Language Dataset},
  journal={arXiv preprint arXiv:2301.11932},
  year = {2023},
}

@inproceedings{franken2013automatic,
  title={Automatic Egyptian hieroglyph recognition by retrieving images as texts},
  author={Franken, Morris and van Gemert, Jan C},
  booktitle=ACMMM,
  year={2013}
}

@inproceedings{assran2023self,
  title={Self-supervised learning from images with a joint-embedding predictive architecture},
  author={Assran, Mahmoud and Duval, Quentin and Misra, Ishan and Bojanowski, Piotr and Vincent, Pascal and Rabbat, Michael and LeCun, Yann and Ballas, Nicolas},
  booktitle=CVPR,
  year={2023}
}

@inproceedings{fifty2023context,
  title={Context-Aware Meta-Learning},
  author={Fifty, Christopher and Duan, Dennis and Junkins, Ronald Guenther and Amid, Ehsan and Leskovec, Jure and Re, Christopher and Thrun, Sebastian},
  year={2024},
  booktitle=ICLR
}

@inproceedings{chi2025learning,
title={Learning to Adapt Frozen {CLIP} for Few-Shot Test-Time Domain Adaptation},
author={Zhixiang Chi and Li Gu and Huan Liu and Ziqiang Wang and Yanan Wu and Yang Wang and Konstantinos N Plataniotis},
booktitle=ICLR,
year={2025},
}

@inproceedings{Yang_2025_Verbalized,
    author    = {Yang, Cheng-Fu and Yin, Da and Hu, Wenbo and Ji, Heng and Peng, Nanyun and Zhou, Bolei and Chang, Kai-Wei},
    title     = {Verbalized Representation Learning for Interpretable Few-Shot Generalization},
    booktitle = ICCV,
    year      = {2025},
}

@inproceedings{Li_2018_ECCV,
    author = {Li, Yingwei and Li, Yi and Vasconcelos, Nuno},
    title = {RESOUND: Towards Action Recognition without Representation Bias},
    booktitle = ECCV,
    year = {2018},
}

@inproceedings{snell2017prototypical,
  title={Prototypical networks for few-shot learning},
  author={Snell, Jake and Swersky, Kevin and Zemel, Richard},
  booktitle=NIPS,
  year={2017}
}

@inproceedings{finn2017model,
  title={Model-agnostic meta-learning for fast adaptation of deep networks},
  author={Finn, Chelsea and Abbeel, Pieter and Levine, Sergey},
  booktitle=ICML,
  year={2017},
}

@inproceedings{sung2018learning,
  title={Learning to compare: Relation network for few-shot learning},
  author={Sung, Flood and Yang, Yongxin and Zhang, Li and Xiang, Tao and Torr, Philip HS and Hospedales, Timothy M},
  booktitle=CVPR,
  year={2018}
}

@inproceedings{ravi2017optimization,
  title={Optimization as a model for few-shot learning},
  author={Ravi, Sachin and Larochelle, Hugo},
  booktitle=ICLR,
  year={2017}
}

@inproceedings{tian2020rethinking,
  title={Rethinking few-shot image classification: a good embedding is all you need?},
  author={Tian, Yonglong and Wang, Yue and Krishnan, Dilip and Tenenbaum, Joshua B and Isola, Phillip},
  booktitle=ECCV,
  year={2020},
}

@inproceedings{schuhmann2022laion,
  title={Laion-5b: An open large-scale dataset for training next generation image-text models},
  author={Schuhmann, Christoph and Beaumont, Romain and Vencu, Richard and Gordon, Cade and Wightman, Ross and Cherti, Mehdi and Coombes, Theo and Katta, Aarush and Mullis, Clayton and Wortsman, Mitchell and others},
  booktitle=NIPS,
  year={2022}
}

@article{dosovitskiy2020image,
  title={An image is worth 16x16 words: Transformers for image recognition at scale},
  author={Dosovitskiy, Alexey and Beyer, Lucas and Kolesnikov, Alexander and Weissenborn, Dirk and Zhai, Xiaohua and Unterthiner, Thomas and Dehghani, Mostafa and Minderer, Matthias and Heigold, Georg and Gelly, Sylvain and others},
  journal={arXiv preprint arXiv:2010.11929},
  year={2020}
}

@inproceedings{kravets2025rethinking,
  title={Rethinking Few Shot CLIP Benchmarks: A Critical Analysis in the Inductive Setting},
  author={Kravets, Alexey and Chen, Da and Namboodiri, Vinay P},
  booktitle=ICCV,
  year={2025}
}

@article{li2024llava,
  title={Llava-onevision: Easy visual task transfer},
  author={Li, Bo and Zhang, Yuanhan and Guo, Dong and Zhang, Renrui and Li, Feng and Zhang, Hao and Zhang, Kaichen and Zhang, Peiyuan and Li, Yanwei and Liu, Ziwei and others},
  journal={arXiv preprint arXiv:2408.03326},
  year={2024}
}

@inproceedings{Kukleva_2021_ICCV,
    author    = {Kukleva, Anna and Kuehne, Hilde and Schiele, Bernt},
    title     = {Generalized and Incremental Few-Shot Learning by Explicit Learning and Calibration Without Forgetting},
    booktitle = ICCV,
    year      = {2021},
}

@manual{nvidia2022h100,
  title        = {NVIDIA H100 Tensor Core GPU Architecture},
  author       = {NVIDIA},
  year         = {2022},
  note         = {Whitepaper}
}
